\documentclass[runningheads]{llncs}
\usepackage{graphicx}
\usepackage{amsmath,amssymb} 
\usepackage{color}

\usepackage{colortbl}
\usepackage{microtype}
\usepackage{amsmath}
\usepackage{amsfonts}
\usepackage{subfigure}
\usepackage{booktabs} 
\usepackage{fixltx2e}
\usepackage{pbox}
\usepackage{multirow}
\usepackage{pifont}
\usepackage{xcolor}
\usepackage{algorithm}
\usepackage{algpseudocode}
\usepackage{tabularx}
\usepackage{wrapfig}
\usepackage{pdfpages}

\usepackage[width=122mm,left=12mm,paperwidth=146mm,height=193mm,top=12mm,paperheight=217mm]{geometry}

\definecolor{bluegray}{rgb}{0.4, 0.6, 0.8}
\definecolor{brightturquoise}{rgb}{0.03, 0.91, 0.87}
\definecolor{cadetblue}{rgb}{0.37, 0.62, 0.63}
\definecolor{NVblue}{rgb}{0.07, 0.12, 0.83}

\definecolor{BUred}{rgb}{0.8, 0.0, 0.0}

\definecolor{caribbeangreen}{rgb}{0.0, 0.8, 0.6}
\definecolor{darkpastelgreen}{rgb}{0.01, 0.75, 0.24}
\definecolor{emerald}{rgb}{0.31, 0.78, 0.47}

\definecolor{ourlightgray}{rgb}{0.9, 0.9, 0.9}
\definecolor{ourdarkgray}{rgb}{0.77, 0.77, 0.77}

\usepackage[pagebackref=true,breaklinks=true,colorlinks=true,citecolor=NVblue,linkcolor=BUred,bookmarks=false]{hyperref}

\usepackage{array}
\usepackage{makecell}

\newcommand{\cmark}{\color{emerald}\ding{52}}%
\newcommand{\xmark}{\color{BUred}\ding{55}}%
\newcommand{\etal}{\textit{et al}.} 
\newcommand{\ie}{\textit{i}.\textit{e}. }
\newcommand{\eg}{\textit{e}.\textit{g}. }

\DeclareMathOperator*{\argmax}{\mathop{arg\,max}}
\DeclareMathOperator*{\argmin}{\mathop{arg\,min}}

\DeclareMathOperator*{\expectation}{\mathop{\mathbb{E}}}

\usepackage{array}
\newcolumntype{P}[1]{>{\centering\arraybackslash}p{#1}}

\newcommand\blfootnote[1]{%
  \begingroup
  \renewcommand\thefootnote{}\footnote{#1}%
  \addtocounter{footnote}{-1}%
  \endgroup
}

\begin{document}

\pagestyle{headings}
\mainmatter

\def\ECCVSubNumber{1769}  

\title{Class-Incremental Domain Adaptation} 

\titlerunning{Class-Incremental Domain Adaptation}

\author{Jogendra Nath Kundu* \and
Rahul Mysore Venkatesh* \and
Naveen Venkat \and \\ Ambareesh Revanur \and R. Venkatesh Babu}


\authorrunning{Kundu \etal}

\institute{Video Analytics Lab, Indian Institute of Science, Bangalore}

\maketitle

\begin{abstract}
We introduce a practical Domain Adaptation (DA) paradigm called Class-Incremental Domain Adaptation (CIDA). Existing DA methods tackle domain-shift but are unsuitable for learning novel target-domain classes. Meanwhile, class-incremental (CI) methods enable learning of new classes in absence of source training data, but fail under a domain-shift without labeled supervision. In this work, we effectively identify the limitations of these approaches in the CIDA paradigm. Motivated by theoretical and empirical observations, we propose an effective method, inspired by prototypical networks, that enables classification of target samples into both shared and novel (one-shot) target classes, even under a domain-shift. Our approach yields superior performance as compared to both DA and CI methods in the CIDA paradigm.\blfootnote{*Equal contribution.}
\end{abstract}

\section{Introduction}
\label{intro}

Deep models have been shown to outperform human evaluators on image recognition tasks~\cite{he2015delving}. However, a common assumption in such evaluations is that the training and the test data distributions are alike.
In the presence of a larger domain-shift~\cite{office} between the training and the test domains, the performance of deep models degrades drastically resulting from the domain-bias~\cite{khosla2012undoingbias,torralba2011unbiased}. Moreover, the recognition capabilities of such models is limited to the set of learned categories, which further limits their generalizability.
Thus, once a model is trained on a source training dataset (the \textit{source} domain), it is essential to further upgrade the model to perform well in the test environment (the \textit{target} domain).

\begin{figure}[!h]
    \centering
    \includegraphics[width=1.0\linewidth]{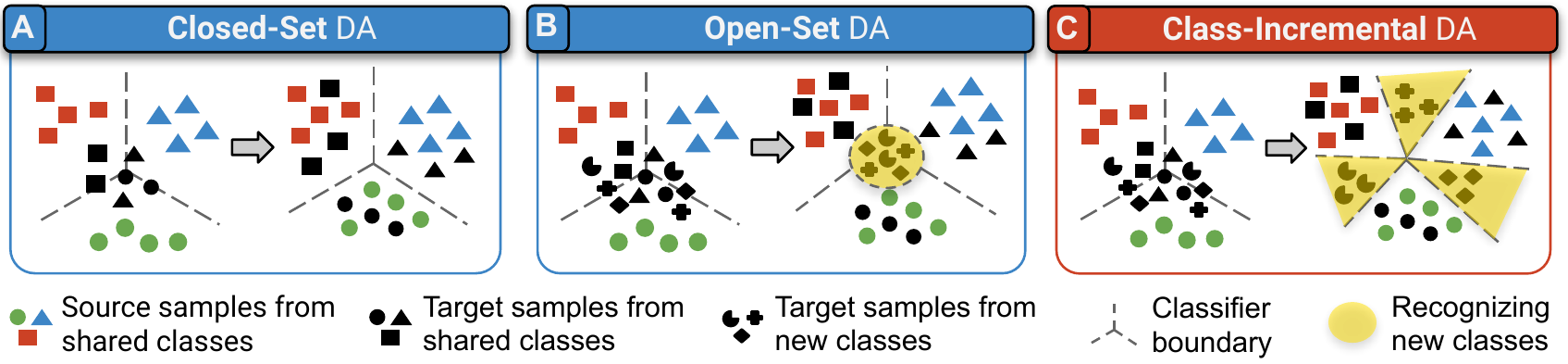}
    \caption{
    {\small
    \textbf{Problem Setting.} \textbf{A)} Closed-set DA assumes a shared label-set between the source and the target domains. \textbf{B)} Open-set DA rejects target samples from unseen categories into a single \textit{unknown} class. \textbf{C)} In Class-Incremental DA, we aim to recognize both shared and new target classes by assigning a unique semantic label to each class.
    }
    }
    \label{fig:Concept}
\end{figure}

For example, consider a \textit{self-driving car} installed with an \textit{object recognition model} trained on urban scenes. Such a model will underperform in rural landscapes (test environment) where objects differ in their visual appearance and the surrounding context.
Moreover, the model will also misclassify objects from unseen categories (\textit{a.k.a} target-private categories) into one of the learned classes. This is a direct result of the domain-shift between urban and rural environments.
A naive approach to address this problem would be to fine-tune~\cite{lwf} the model on an annotated dataset drawn from the target environment. However, this is often not a practical solution 
as acquiring label-rich data is an expensive process. Moreover, 
for an efficient model upgrade, it is also imperative that the model supports adaptation to new domains and tasks, without re-training on the source training data \cite{lwm,lwf} from scratch.
Motivated by these challenges, in this paper we ask ``how to effectively upgrade a trained model to the target domain?''.

In the literature, this question has been long-standing. A line of work called Unsupervised Domain Adaptation (UDA) \cite{ben2010theory,ben2007analysis,chang2019dsbn,kundu2019adapt,kuroki2019unsupervised_discrepancy,Adapt3long2016unsupervised,nath2018adadepth,tzeng2017adversarial} has emerged that offers an elegant solution to the domain-shift problem. 
In UDA, the usual practice~\cite{NormalAdaptfirstBackprop,mada} is to obtain a labeled source dataset and unlabeled targets samples, to perform adaptation under the co-existence of samples from both the domains.
However, most UDA methods~\cite{NormalAdaptfirstBackprop,Adapt4gong2012geodesic,saito2018maximum,NormalAdapttzeng2014deep} assume that the two domains share the same label space (as shown in Fig.~\ref{fig:Concept}{\color{BUred}A}), making them impractical in real-world where a target domain potentially contains unseen categories (\textit{in the self-driving car example, novel objects occur in the deployed environment}).
To this end, open-set DA~\cite{OpenSetbaktashmotlagh2018learning,panareda2017open,inheritune,Saito_2018_ECCV} and universal DA~\cite{usfda,UDA_2019_CVPR} have gained attention, where the target domain is allowed to have novel (target-private) classes not present in the source domain. These target-private samples are assigned an ``\textit{unknown}'' label (see Fig.~\ref{fig:Concept}{\color{BUred}B}). As a result, target-private samples with diverse semantic content get clustered together in a single ``\textit{unknown}'' class in the latent space.

While UDA methods tackle the domain-shift problem, these require simultaneous access to both source and target domain samples, which makes them unsuitable in cases where the source training data is proprietary~\cite{inheritune,dfkd,zskd} (\textit{e.g. in a self-driving car}), or simply unavailable during model upgrade~\cite{lwm,usfda,lwf}. Moreover, these methods can only detect new target categories as a single \textit{unknown} class~\cite{panareda2017open}, and cannot assign individual semantic labels to such categories (Fig. \ref{fig:Concept}{\color{BUred}C}). Thus, these methods do not truly facilitate model upgrade (\textit{e.g. adding new classes to the recognition model}) thereby having a limited practical use-case.

Another line of work consists of Class-Incremental (CI) learning methods \cite{IncrementalCastro_2018_ECCV,lwf,peng2017incrementally,ruping2001incremental,wu2019largescaleincrementallearning} which aim at adding new classes to a trained model while preserving the performance on the previously learned classes.
Certain methods \cite{lwm} achieve this even without accessing the source training data (hereon, we call such methods as \textit{source-free}). However, these methods are not tailored to address domain-shift (\textit{thus, in our example, the object recognition model would still underperform in rural scenarios}). Moreover, many of these methods \cite{IncrementalCastro_2018_ECCV,lwm,icarl} require the target data to be labeled, which is impractical for real world applications.

\begin{wraptable}{r}{0.44\textwidth}
\caption{
{Characteristic comparison based on the support for \textit{source-free} (\textbf{SF}), class-incremental (\textbf{CI}) model upgrade under domain-shift (\textbf{DA}).}
}
\centering
\label{tab:characteristic_comparison}
\setlength\tabcolsep{17pt}
\renewcommand{\arraystretch}{1.1}
\resizebox{0.44\columnwidth}{!}{
\begin{tabular}{|l|c|c|c|}
\hline

{\makecell{\textbf{Method}}} & {\makecell{\textbf{SF}}} & {\makecell{\textbf{CI}}} & {\makecell{\textbf{DA}}} \\
\hline
DANN~\cite{ganin2016domain} & \xmark & \xmark & \cmark \\
OSBP~\cite{Saito_2018_ECCV} & \xmark & \xmark & \cmark \\
UAN~\cite{UDA_2019_CVPR} & \xmark & \xmark & \cmark \\
STA~\cite{sta_open_set} & \xmark & \xmark & \cmark \\
\hline
LETR~\cite{feifei-incremental} & \xmark & \cmark & \cmark  \\
E2E~\cite{IncrementalCastro_2018_ECCV} & \xmark & \cmark & \xmark  \\
iCaRL~\cite{icarl} & \xmark & \cmark & \xmark  \\
LwF-MC~\cite{lwm} & \cmark & \cmark & \xmark  \\
LwM~\cite{lwm} & \cmark & \cmark & \xmark  \\
\hline
Ours & \cmark & \cmark & \cmark  \\
\hline

\end{tabular}
}
\end{wraptable}

To summarize, UDA and CI methods address different challenges under separate contexts and neither of them alone suffices practical scenarios. A characteristic comparison against prior arts is given in Table~\ref{tab:characteristic_comparison}. Acknowledging this gap between the available solutions and their practical usability, in this work we introduce a new paradigm called Class-Incremental Domain Adaptation (CIDA) with the best of both worlds. While formalizing the paradigm, we draw motivation from both UDA and CI and address their limitations in CIDA.

In CIDA, we aim to adapt a source-trained model to the desired target domain in the presence of domain-shift as well as unseen classes using a minimal amount of labeled data.
To this end, we propose a novel training strategy which enables a \textit{source-free} upgrade to an unlabeled target domain by utilizing one-shot target-private samples. Our approach is motivated by prototypical networks~\cite{snell2017prototypical} which exhibit a simpler inductive bias in the limited data regime. We now review the prior arts and identify their limitations to design a suitable approach for CIDA. Our contributions are as follows:

\begin{itemize}

\item We formalize a novel Domain Adaptation paradigm, Class-Incremental Domain Adaptation (CIDA), which enables the recognition of both shared and novel target categories under a domain-shift. 

\item We discuss the limitations of existing approaches and identify the challenges involved in CIDA to propose an effective training strategy for CIDA.

\item The proposed solution is motivated by theoretical and empirical observations and outperforms both UDA and CI approaches in CIDA.

\end{itemize}

\section{Background}
\label{sec:background}

Before formalizing the CIDA paradigm, we review the prior methods and study their limitations. In the UDA problem, we consider a labeled source domain with the label-set $\mathcal{C}_s$ and an unlabeled target domain with the label-set $\mathcal{C}_t$. The goal is to improve the task-performance on the target domain by transferring the task-relevant knowledge from the source to the target domain.

The most popular UDA approach \cite{ganin2016domain,kuroki2019unsupervised_discrepancy,Adapt2long2015learning,NormalAdaptsankaranarayanan2018generate,coral,NormalAdapttzeng2014deep} is to learn a predictor $h(x) = g \circ f(x)$ having a domain-agnostic feature extractor $f$ that is common to both the domains, and a classifier $g$ which can be learned using source supervision. These methods align the latent features $f(\cdot)$ of the two domains and use the classifier $g$ to predict labels for the target samples. A theoretical upper bound \cite{ben2010theory} for the target-domain risk of such predictors is as follows,

\begin{equation}
\label{upper_bound}
    \epsilon_t(g) \leq \epsilon_s(g) + \frac{1}{2} d_{\mathcal{H} \Delta \mathcal{H}}(s, t) + \lambda
\end{equation}

where, given a hypothesis space $\mathcal{H}$, $\epsilon_s$ and $\epsilon_t$ denote the expected risk of the classifier $g \in \mathcal{H}$ in the source and the target domains respectively, and $d_{\mathcal{H} \Delta \mathcal{H}} = 2 \operatorname{sup}_{g, g^\prime \in \mathcal{H}} |\epsilon_s(g,g^\prime) - \epsilon_t(g,g^\prime)|$ measures the distribution shift (or the domain discrepancy) between the two domains and $\lambda = \min_{g \in \mathcal{H}} \epsilon_s(g) + \epsilon_t(g)$ is a constant that measures the risk of the optimal joint classifier.

Notably, UDA methods aim to minimize the upper bound of the target risk (Eq.~\ref{upper_bound}) by minimizing the distribution shift $d_{\mathcal{H}\Delta\mathcal{H}}$ in the latent space $f(\cdot)$, while preserving a low source risk $\epsilon_s$. This works well under the closed-set assumption (\ie $\mathcal{C}_s = \mathcal{C}_t$). However, in the presence of target-private samples (\ie samples from $\mathcal{C}_t^{\prime} = \mathcal{C}_t \setminus \mathcal{C}_s$), a direct enforcement of such constraints often degrades the performance of the model, even on the shared categories - a phenomenon known as negative transfer \cite{pan2009survey}. This is due to two factors. Firstly, a shared feature extractor ($f$), which is expected to generalize across two domains, acts as a bottleneck to the performance on the target domain. 
Secondly, a shared feature extractor enforces a common semantic granularity in the latent space ($f(\cdot)$) across both the domains. This is especially unfavorable in CIDA, where the semantic space must be modified to accommodate target-private categories (see Fig.~\ref{fig:cluster_transit}).

\textbf{Why are UDA methods insufficient?} Certain UDA methods \cite{sta_open_set,UDA_2019_CVPR} tackle negative transfer by detecting the presence of target-private samples and discarding them during domain alignment. As a result, these samples (with diverse semantic content) get clustered into a single \textit{unknown} category. While this improves the performance on the shared classes, it disturbs the semantic granularity of the latent space (\ie $f(\cdot)$), making the model unsuitable for a class-incremental upgrade. This additional issue must be tackled in CIDA. 

To demonstrate this effect, 
we employ the state-of-the-art open-set DA method STA~\cite{sta_open_set} for image recognition on the 
Amazon $\rightarrow$ DSLR task of Office~\cite{office} dataset. A possible way to extend STA for CIDA would be to
collect the target samples that are predicted as \textit{unknown} (after adaptation) and obtain few-shot labeled samples from this set (by randomly labeling, say, $5\%$ of the samples). One could then train a classifier using these labeled samples. We follow this approach and over 5 separate runs, we calculate the class-averaged accuracy. The model achieves an accuracy of $95.9\pm0.3 \%$ on the shared classes, while only $17.7\pm3.5 \%$ on the target-private classes. See Suppl. for experimental details. This clearly indicates that the adaptation disturbs the granularity of the semantic space~\cite{kundu2019gantree}, which is no more useful for discriminating among novel target categories.

\textbf{Why are CI methods insufficient?} Works such as \cite{IncrementalCastro_2018_ECCV,feifei-incremental,icarl} use an exemplary set to receive supervision for the source classes $\mathcal{C}_s$ along with labeled samples from target-private classes. \cite{feifei-incremental} aims to address domain-shift using labeled samples. However, the requirement of the source data during model upgrade is a severe drawback for practical applications~\cite{lwf}. While \cite{lwm} is \textit{source-free}, it still assumes access to labeled target samples, which may not be viable in practical deployment scenarios. As we show in Sec.~\ref{sec:experiments}, these methods yield suboptimal results in the presence of limited labeled data. Nevertheless, most CI methods are not geared to tackle domain-shift. Thus, the assumption that the source-model is proficient in classifying samples in $\mathcal{C}_s$ \cite{lwm}, will not hold good for the target domain. To the best of our knowledge, the most closely related CI work is \cite{dong2018_oneshot_domainadaptation} that uses a reinforcement-learning based framework to select source samples during one-shot learning. However, \cite{dong2018_oneshot_domainadaptation} assumes non-overlapping label sets ($\mathcal{C}_s \cap \mathcal{C}_t = \phi$), and does not consider the effect of negative transfer during model upgrade.

\textbf{Why do we need CIDA?} Prior arts independently address the problem of class-incremental learning and unsupervised adaptation in seperate contexts, by employing learning procedures specific to the problem at hand. As a result of this specificity, they are not equipped to address practical scenarios (\textit{such as the self-driving car example} in Sec.~\ref{intro}). Acknowledging their limitations, we propose CIDA where the focus is to improve the performance on the target domain to achieve class-incremental recognition in the presence of domain-shift. This makes CIDA more practical and more challenging than the available DA paradigms.

\textbf{What do we assume in CIDA?} To realize a concrete solution, we make the following assumptions that are within the bounds of a practical DA setup. Firstly, considering that the labeled source dataset may not be readily available to perform a model upgrade, we consider the \textit{adaptation step to be source-free}. Accordingly, we propose an effective source-model training strategy which allows \textit{source-free} adaptation to be implemented in practice. Secondly, as the target domain may be label-deficient, we pose CIDA as an Unsupervised DA problem wherein the \textit{target samples are unlabeled}. However, conceding that it may be impractical to discover semantics for unseen target classes in a completely unsupervised fashion, we assume that we can obtain a single labeled target sample for each target-private class $\mathcal{C}_t^{\prime}$ (\textit{one-shot target-private} samples). This can be perceived as the knowledge of new target classes that must be added during the model upgrade. Finally, the overarching objective in CIDA is to \textit{improve the performance in the target domain} while the performance on the source domain remains secondary.

The assumptions stated above can be interpreted as follows. In CIDA, we first quantify the upgrade that is to be performed. We identify ``what domain-shift is to be tackled?'' by collecting unlabeled target domain samples, and determine ``what new classes are to be added?'' by obtaining one-shot target-private samples. This deterministic quantification makes CIDA different from UDA and CI methods, and enhances the reliability of a \textit{source-free} adaptation algorithm. In the next section, we formalize CIDA and describe our approach to solve the problem.

\section{Class-Incremental Domain Adaptation}
\label{sec:Our Approach}

 Let $\mathcal{X}$ and $\mathcal{Y}$ be the input and the label spaces. The source and the target domains are characterized by the distributions $p$ and $q$ on $\mathcal{X} \times \mathcal{Y}$. We denote the set of labeled source samples as $\mathcal{D}_s = \{(\mathbf{x}_s, y_s) : (\mathbf{x}_s, y_s) \sim p\}$ with label set $\mathcal{C}_s$ and the set of unlabeled target samples as $\mathcal{D}_t = \{\mathbf{x}_t: \mathbf{x}_t \sim q_{\mathcal{X}}\}$ with label-set $\mathcal{C}_t$, where $q_{\mathcal{X}}$ denotes the marginal input distribution and $\mathcal{C}_s \subset \mathcal{C}_t$. The set of target-private classes is denoted as $\mathcal{C}_t^{\prime} = \mathcal{C}_t \setminus \mathcal{C}_s$.
See Suppl. for a notation table.
 To perform class-incremental upgrade, we are given one target sample from each target-private category $\{(\mathbf{\tilde{x}}_t^{c}, \tilde y_t^{c})\}_{c \in \mathcal{C}_t^{\prime}}$ (one-shot target-private samples). Further, we assume that source
 samples are unavailable during model upgrade~\cite{usfda,inheritune,lwf}.
 Thus, the goal is to train a model on the source domain, and later, upgrade the model (address domain-shift and learn new classes) for the target domain. Accordingly, we formalize a two-stage approach as follows,

\begin{enumerate}
    \item \textbf{Foresighted source-model training.} It is imperative that a source-trained model supports \textit{source-free} adaptation. 
    Thus, 
    during source training,
    we aim to suppress the domain and category bias~\cite{khosla2012undoingbias} that culminates from
    overconfident class-predictions. Specifically, we augment the model with the capability of out-of-distribution~\cite{lee2018training_ood} detection.
    This step is inspired by prototypical networks that have a simpler inductive bias in the limited data regime~\cite{snell2017prototypical}. Finally, the source-model is shipped along with prototypes as meta-data, for performing a future \textit{source-free} upgrade.

    \item \textbf{Class-Incremental DA.} During CIDA,
    we aim to align the target samples from shared classes with the high-source-density regions in the latent space, allowing the reuse of the source classifier. Further, we must accommodate new target classes in the latent space while preserving the semantic granularity.
    We achieve both these objectives by learning a target-specific latent space in which we obtain learnable centroids called \textit{guides}
    that are used to gradually steer the target features into separate clusters. We theoretically argue and empirically verify that this enables a suitable ground for CIDA.
\end{enumerate}

\subsection{Foresighted source-model training}
\label{sec:source_training}

\begin{figure}[t]
    \centering
    \includegraphics[width=1\linewidth]{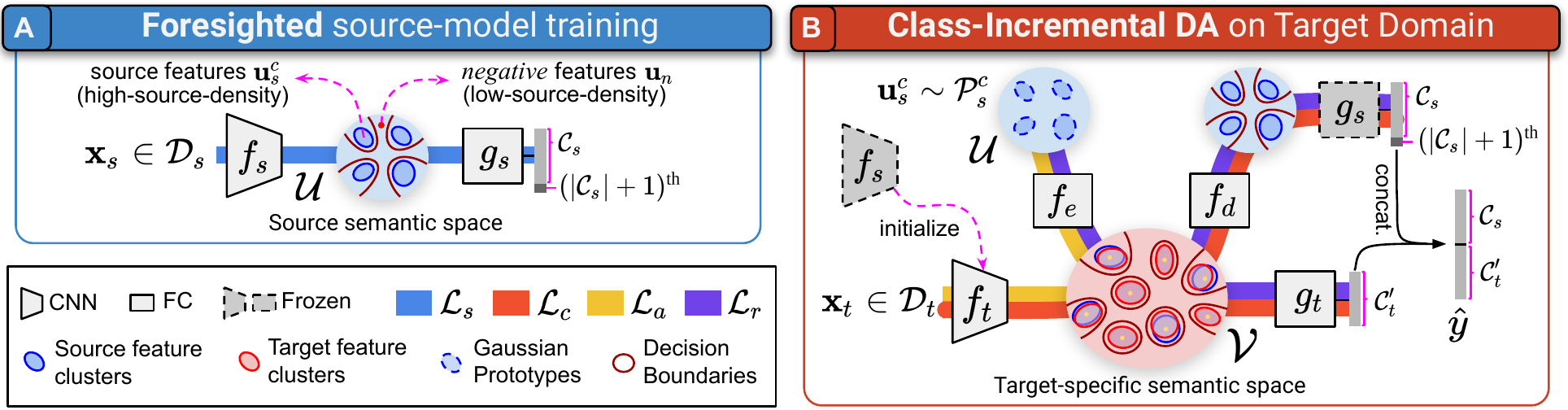}
    \caption{
    {
    \textbf{Our Approach.} \textbf{A)} The source-model is trained with an additional $(|\mathcal{C}_s|+1)^{\text{th}}$ class representing out-of-distribution (OOD) region.
    \textbf{B)} During CIDA, we learn a target-specific feature extractor $f_t$ (to minimize domain-shift) and classifier $g_t$ (to learn $\mathcal{C}_t^{\prime}$).
    The adaptation process aligns the shared classes, and separates the target-private classes.
    Colored lines represent the gradient pathway for each loss.
    }
    }
    \label{fig:approach}
\end{figure}

The architecture for the source model contains a feature extractor $f_s$ 
and a $(|\mathcal{C}_s|+1)$-class classifier $g_s$ (see Fig.~\ref{fig:approach}{\color{BUred}A}). We denote the latent-space by $\mathcal{U}$.
A naive approach to train the source-model would be using the cross-entropy loss,

\begin{equation}
\label{eq:ce_loss}
    \mathcal{L}_{vanilla} : \expectation_{(\mathbf{x}_s, y_s) \sim p} l_{ce}\left(g_s \circ f_s (\mathbf{x}_s), y_s\right)
\end{equation}

 where, $\circ$ denotes composition. However, enforcing $\mathcal{L}_{vanilla}$ alone 
 biases the model towards source domain characteristics. As a result, the model learns highly discriminative features and mis-classifies out-of-distribution samples into one of the learned categories ($\mathcal{C}_s$) with high confidence~\cite{inheritune}. For \textit{e.g.}, an MNIST image classifier is shown to yield a predicted class-probability of 91$\%$ on random input~\cite{hendrycks2017baseline}. We argue that such effects are due to the domain and category bias culminating from the overconfident predictions.
Thus, we aim to suppress this bias in the presence of the source samples for a reliable \textit{source-free} upgrade.

We note two requirements for a source-trained model suitable for CIDA. First, we must penalize overconfident predictions \cite{pereyra2017regularizing} which is a crucial step to enable generalization over unseen target categories. This will aid in mitigating the effect of negative-transfer (discussed in Sec.~\ref{sec:background}). 
Second, we aim for \textit{source-free} adaptation in CIDA, which calls for an alternative to source samples.
We satisfy both these requirements using class-specific gaussian prototypes~\cite{fort2017gaussian_prototypical_network,snell2017prototypical} as follows.

\textbf{a) Gaussian Prototypes.} We define a Gaussian Prototype for a class $c$ as $\mathcal{P}_s^c = \mathcal{N}(\boldsymbol{\mu}_s^c, \boldsymbol{\Sigma}_s^c)$ where $\boldsymbol{\mu}_s^c$ and $\boldsymbol{\Sigma}_s^c$ are the 
mean and the 
covariance obtained over the features $f(\mathbf{x}_s)$ for samples $\mathbf{x}_s$ in class $c$.
In other words, a Gaussian Prototype is a multivariate Gaussian prior defined for each class
in the latent space $\mathcal{U}$. Similarly, a global Gaussian Prototype is defined as $\mathcal{P}_s = \mathcal{N}(\boldsymbol{\mu}_s, \boldsymbol{\Sigma}_s)$, where $\boldsymbol{\mu}_s$ and $\boldsymbol{\Sigma}_s$ are calculated over the features $f_s(\mathbf{x}_s)$ for all source samples $\mathcal{D}_s$. We hypothesize that at the $\mathcal{U}$-space, we can approximate the class semantics using these Gaussian priors which can be leveraged for \textit{source-free} adaptation. 

To ensure that this Gaussian approximation 
is accurate, 
we explicitly enforce the source features to attain a higher affinity towards these class-specific Gaussian priors. We refer to this as the \textit{class separability} objective defined as,

\begin{equation}
    \mathcal{L}_{s1} : \expectation_{(\mathbf{x}_s, y_s) \sim p} 
    -\operatorname{log} \Big( ~\exp{(\mathcal{P}_s^{y_s}(\mathbf{u}_s))} ~/ \sum_{c \in \mathcal{C}_s} \exp{(\mathcal{P}_s^{c}(\mathbf{u}_s))}
    ~\Big)
\end{equation}

where $\mathbf{u}_s = f(\mathbf{x}_s)$, and the term inside the $\operatorname{logarithm}$ is the posterior probability of a feature $\mathbf{u}_s$ corresponding to its class $y_s$ (obtained as the softmax over likelihoods $\mathcal{P}_s^c(\mathbf{u}_s)$). 
In effect, $\mathcal{L}_{s1}$ drives the latent space to form well-separated, compact clusters for each class $c \in \mathcal{C}_s$. 
We verify in Sec.~\ref{sec:experiments} that compact clusters enhance the reliability of a \textit{source-free} model upgrade, where the clusters must rearrange to attain a semantic granularity suitable for the target domain.

\textbf{b) Negative Training.} While $\mathcal{L}_{s1}$ enforces well-separated feature clusters, it does not ensure tight decision boundaries, without which
the classifier
misclassifies out-of-distribution (OOD) samples~\cite{lee2018training_ood} with high confidence. This overconfidence issue must be resolved to effectively learn new target categories.
Certain prior works \cite{intrinsic-adversarial} suggest that a Gaussian Mixture Model based likelihood threshold could effectively detect OOD samples. We argue that
additionally, the classifier $g_s$ should also be capable of assigning a low confidence to OOD samples \cite{lee2018training_ood}, forming tight decision boundaries around the source clusters (as in Fig.~\ref{fig:approach}{\color{BUred}A}). 

We leverage the Gaussian Prototypes to generate \textit{negative} feature samples to model the low-source-density (OOD) region.
The \textit{negative} samples are denoted as $\mathcal{D}_n = \{ (\mathbf{u}_n, y_n) : (\mathbf{u}_n, y_n) \sim r\}$ where $r$ is the distribution of the OOD regime. More specifically, we obtain the samples $\mathbf{u}_n$ from the global Gaussian Prototype $\mathcal{P}_s$ which are beyond 3-$\sigma$ confidence interval of all class-specific Gaussian Prototypes $\mathcal{P}_s^c$ (see Suppl. for an algorithm). These \textit{negative} samples correspond to the $(|\mathcal{C}_s|+1)^{\text{th}}$ category and the classifier $g_s$ is trained to assign a low confidence to such samples (see Fig.~\ref{fig:approach}{\color{BUred}A}). Thus, the cross-entropy loss in Eq.~\ref{eq:ce_loss} is modified as:

\begin{equation}
    \mathcal{L}_{s2} : \expectation_{ (\mathbf{x}_s, y_s) \sim p } l_{ce}(g_s \circ f_s (\mathbf{x}_s), y_s) ~~ +
    \expectation_{(\mathbf{u}_n, y_n) \sim r} 
    l_{ce}(g_s (\mathbf{u}_n), y_n)
\end{equation}

By virtue of $\mathcal{L}_{s2}$, the classifier $g_s$ assigns a high source-class confidence to samples in $\mathcal{D}_s$, and a low source-class confidence 
to samples in $\mathcal{D}_n$. 
Thus, $g_s$ learns compact decision boundaries (as shown in Fig.~\ref{fig:approach}{\color{BUred}A}).

\textbf{c) Optimization.} We train $\{f_s, g_s\}$ via alternate minimization of $\mathcal{L}_{s1}$ and $\mathcal{L}_{s2}$ using Adam~\cite{adam} optimizers (see Suppl.). Effectively, the total loss $\mathcal{L}_s = \mathcal{L}_{s1} + \mathcal{L}_{s2}$ enforces the Cluster Assumption at the $\mathcal{U}$-space (via $\mathcal{L}_{s1}$) that enhances the model's generalizability~\cite{chapelle2005semi_cluster_assumption,cluster_assumption_grandvalet2005semi}, and, mitigates the overconfidence issue (via $\mathcal{L}_{s2}$) thereby reducing the discriminative bias towards the source domain.
We update the Gaussian Prototypes and the \textit{negative} samples at the end of each epoch.
Once trained, the source-model is ready to be shipped along with the Gaussian Prototypes as meta-data. Note, in contrast to source data, Gaussian Prototypes are cheap and can be readily shared (similar to BatchNorm~\cite{ioffe2015batchnorm} statistics).

\subsection{Class-Incremental DA on the Target Domain}
\label{sec:incremental_adaptation}
 
Following the CIDA paradigm during the model upgrade, we have access to a source model $\{f_s, g_s\}$ and its meta-data (Gaussian Prototypes $\mathcal{P}_s^c$), unlabeled target samples $\mathcal{D}_t$, and one-shot target-private samples $\{(\tilde{\mathbf{x}}_t^c, \tilde{y}_t^c)\}_{c \in \mathcal{C}_t^{\prime}}$. We now formalize an approach that tightens the target risk bound (Eq.~\ref{upper_bound}) exploiting a foresighted source-model trained using $\mathcal{D}_s \cup \mathcal{D}_n$. Recall that the bound comprises of three terms - source risk ($\epsilon_s$), distribution shift ($d_{\mathcal{H}\Delta\mathcal{H}}$) and the constant $\lambda$.

\textbf{a) Learning target features.} 
A popular strategy for UDA is to learn domain-agnostic features~\cite{sta_open_set,Saito_2018_ECCV,shu2019transferable_thuml_weaklysupervised_da}. 
However, as argued in Sec.~\ref{sec:background}, in CIDA we must learn a target-specific latent space (annotated as $\mathcal{V}$ in Fig.~\ref{fig:approach}{\color{BUred}B}) which attains a semantic granularity suitable for the target domain. To this end, we introduce a target-specific feature extractor $f_t$ that is initialized from $f_s$. Informally, this process ``initializes the $\mathcal{V}$-space from the $\mathcal{U}$-space''. Thereafter, we gradually rearrange the feature clusters in the $\mathcal{V}$-space to learn suitable target semantics. To receive stable gradients, $g_s$ is kept frozen throughout adaptation. Further, we introduce a classifier $g_t$ to learn the target-private categories $\mathcal{C}_t^{\prime}$ (see Fig.~\ref{fig:approach}{\color{BUred}B}).

\textbf{b) Domain projection.} 
The key to effectively learn target-specific semantics is to establish a transit mechanism between the $\mathcal{U}$-space (capturing the semantics of the learned classes $\mathcal{C}_s$) and the $\mathcal{V}$-space (where $\mathcal{C}_t$ must be learned). 
We address this using the domain projection networks $f_e : \mathcal{U} \rightarrow \mathcal{V}$ and $f_d : \mathcal{V} \rightarrow \mathcal{U}$. Specifically, we obtain feature samples from the Gaussian Prototypes $\mathbf{u}_s^c \sim \mathcal{P}_s^c$ 
for each class $c \in \mathcal{C}_s$ (called as proxy-source samples). 
Thereafter, we formalize the following losses to minimize the source risk ($\epsilon_s$ in Eq.~\ref{upper_bound}) during adaptation,

\begin{equation}
\label{eq:loss_r}
    \mathcal{L}_{r1} : 
    \expectation_{\mathbf{u}_s^c \sim \mathcal{P}_s^c} ~ l_{ce}(\hat{y}(\mathbf{u}_s^c), c) 
    ~~~~;~~~ 
    \mathcal{L}_{r2} : 
    \expectation_{\mathbf{u}_s^c \sim \mathcal{P}_s^c} l_2(f_d \circ f_e (\mathbf{u}_s^c), \mathbf{u}_s^c)^2
\end{equation}

where $l_2$ is the euclidean distance and the output $\hat{y}(\cdot)$ is the concatenation (Fig.~\ref{fig:approach}{\color{BUred}B}) of logits pertaining to $\mathcal{C}_s$ ($g_s \circ f_d \circ f_e(\mathbf{u}_s^c)|_{c \in \mathcal{C}_s}$) and those of $\mathcal{C}_t$ ($g_t \circ f_e (\mathbf{u}_s^c)$). The total loss $\mathcal{L}_r = \mathcal{L}_{r1} + \mathcal{L}_{r2}$ acts as a regularizer,
where $\mathcal{L}_{r1}$ preserves the semantics of the learned classes in the $\mathcal{V}$-space, while $\mathcal{L}_{r2}$ prevents degenerate solutions. In Sec.~\ref{sec:experiments}, we show that $\mathcal{L}_r$ mitigates catastrophic forgetting~\cite{goodfellow2013empirical_catastrophic_forgetting} (by minimizing $\epsilon_s$ in Eq.~\ref{upper_bound}) that would otherwise occur in a \textit{source-free} scenario.

\textbf{c) Semantic alignment using guides.}
We aim to align target samples from shared classes $\mathcal{C}_s$ with the high source-density region (proxy-source samples) and disperse the target-private samples away into the low source-density region (\ie the \textit{negative} regime). Note, as the source model was trained on $\mathcal{D}_s$ augmented with $\mathcal{D}_n$, this process would entail the minimization of $d_{\mathcal{H}\Delta\mathcal{H}}$ (Eq.~\ref{upper_bound}) measured between the target and the augmented source distributions in the $\mathcal{V}$-space.

\begin{figure*}[t]
    \centering
    \includegraphics[width=0.97\linewidth]{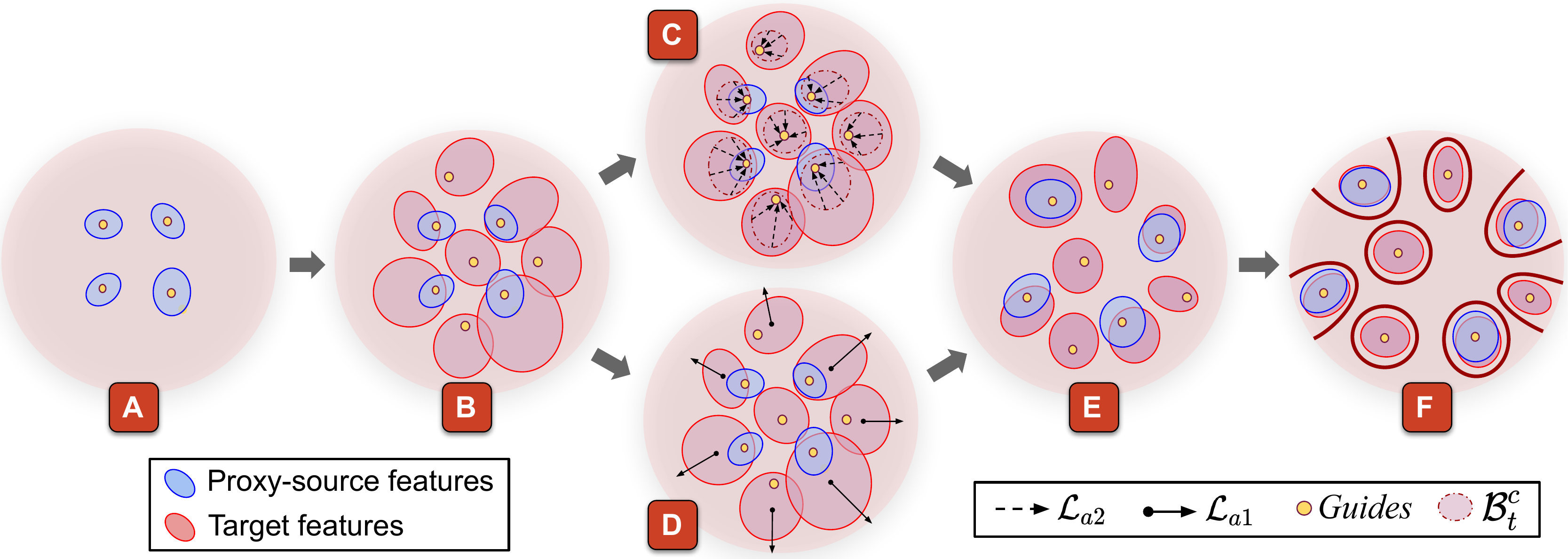}
    \caption{
    {
    \textbf{Semantic Alignment using \textit{guides} in $\mathcal{V}$-space.} \textbf{A)} $\mathcal{V}$ is initialized from $\mathcal{U}$. \textbf{B)} Domain-shift between target and proxy-source features. \textbf{C)} $\mathcal{L}_{a2}$ steers the confident target samples ($\mathcal{B}_t^c$) towards the corresponding \textit{guides} ($\mathbf{v}_g^c$). \textbf{D)} $\mathcal{L}_{a1}$ separates the clusters, making space for target-private classes. \textbf{E)} This rearrangement aligns the shared classes while separating the target-private classes. \textbf{F)} The classifiers $\{g_s, g_t\}$ recognize all target classes by assigning an individual semantic label to each class.
    }
    }
    \label{fig:cluster_transit}
\end{figure*}

To achieve this, we obtain a set of $|\mathcal{C}_t|$ \textit{guides} ($\mathbf{v}_g^c$) that act as representative centers for each class $c \in \mathcal{C}_t$ in the $\mathcal{V}$-space. 
We model the euclidean distance to a \textit{guide} as a measure of class confidence, using which we can assign a pseudo class-label~\cite{lee2013pseudo} to the target samples.
These pseudo-labels can be leveraged to rearrange the target features into separate compact clusters (Fig.~\ref{fig:cluster_transit}{\color{BUred}B-F}).
 Note that $\mathcal{L}_{s1}$ (\textit{class separability} objective) enforced during the source training is crucial to improve the reliability of the \textit{guides} during adaptation.

We consider the features of the one-shot target-private samples $f_t(\tilde{\mathbf{x}}_s^c)$ as the \textit{guides} for the target-private classes. 
Further, since $\mathcal{V}$ is initialized from $\mathcal{U}$, one might consider the source class-means $\boldsymbol{\mu}_s^c$ as the \textit{guides} for the shared classes.
However, we found that a fixed \textit{guide} representation (\eg $\boldsymbol{\mu}_s^c$) hinders the placement of target-private classes.
Thus, we obtain trainable \textit{guides} for the shared classes as $f_e(\boldsymbol{\mu}_s^c)$, by allowing $f_e$ to modify the placement of the \textit{guides} in the $\mathcal{V}$-space (Fig.~\ref{fig:cluster_transit}).
This allows all the \textit{guides} to rearrange and steer the target clusters in the $\mathcal{V}$-space as the training proceeds.
To summarize, the \textit{guides} are computed as $\mathbf{v}_g^{c} = f_e(\boldsymbol{\mu}_s^{c}) ~\forall c \in \mathcal{C}_s$, and, $\mathbf{v}_g^c = f_t (\tilde{\mathbf{x}}_t^c) ~\forall c \in \mathcal{C}_t^{\prime}$.

To minimize $d_{\mathcal{H}\Delta\mathcal{H}}$ (Eq.~\ref{upper_bound}), we must first detect the target-shared and target-private samples and then perform feature alignment. 
To this end, for a target feature $\mathbf{v}_t = f_t(\mathbf{x}_t)$, we obtain the euclidean distance $d$ to its nearest \textit{guide}, and assign a pseudo-label $k$ corresponding to the class represented by the \textit{guide} as, $  d = \min_{c \in \mathcal{C}_t} ~l_2(\mathbf{v}_t, \mathbf{v}_g^c)$, and, $k = \argmin_{c \in \mathcal{C}_t} ~l_2(\mathbf{v}_t, \mathbf{v}_g^c)$.

 Using pseudo-labeled samples we obtain Gaussian Prototypes $\mathcal{P}_t^c = \mathcal{N}(\boldsymbol{\mu}_t^c, \boldsymbol{\Sigma}_t^c)$ $\forall c \in \mathcal{C}_t$ in the $\mathcal{V}$-space (as done in Sec.~\ref{sec:source_training}{\color{BUred}a}), and enforce the \textit{class separability} objective. 
 Further, for each \textit{guide} $\mathbf{v}_g^c$ ($c \in \mathcal{C}_t$), we define a set $\mathcal{B}_t^c$
 of the closest $n$-percent target samples based on the distance $d$ (see Suppl. for the algorithm). Notionally, $\mathcal{B}_t^c$ represents the confident target samples which are then pulled closer to $\mathbf{v}_g^c$. These two losses are defined as,

 \begin{equation}
    \mathcal{L}_{a1} : \expectation_{\mathbf{x}_t \sim q_{\mathcal{X}}}  - \operatorname{log}\Big(\exp{(\mathcal{P}_t^k(\mathbf{v}_t))} / \sum_{c \in \mathcal{C}_t} \exp{(\mathcal{P}_t^c(\mathbf{v}_t))}\Big) ~;~
    \mathcal{L}_{a2} : 
    \expectation_{\mathbf{x}_t \sim \mathcal{B}_t^c} l_2(\mathbf{v}_t, \mathbf{v}_g^c)^2
 \end{equation}
 
 The total adaptation loss is $\mathcal{L}_a = \mathcal{L}_{a1} + \mathcal{L}_{a2}$. Overall, $\mathcal{L}_{a}$ pulls the target-shared samples towards the high source-density region and separates the target-private clusters away from the high source-density regions (Fig.~\ref{fig:cluster_transit}{\color{BUred}B-E}). This results in a superior alignment thereby minimizing $d_{\mathcal{H}\Delta\mathcal{H}}$. Particularly, the separation caused by $\mathcal{L}_{a1}$ minimizes the negative influence of target-private samples during adaptation, thereby preventing negative transfer~\cite{sta_open_set}. $\mathcal{L}_{a2}$ ensures compact feature clusters which aids in preserving the semantic granularity across the target classes.

\textbf{d) Learning target-private classes.} Finally, to learn new target classes, we apply cross-entropy loss on the confident target samples $\mathcal{B}_t^c$ ($c \in \mathcal{C}_t$) as,

 \begin{equation}
 \label{eq:loss_ce_target}
     \mathcal{L}_c : 
     \expectation_{\mathbf{x}_t \sim \mathcal{B}_t^c} l_{ce}(\hat{y}(\mathbf{v}_t), c)
 \end{equation}

 where the output $\hat{y}(\cdot)$ is obtained similar to that in Eq.~\ref{eq:loss_r}, by concatenating the logits $g_s \circ f_d (\mathbf{v}_t)|_{c \in \mathcal{C}_s}$ and $g_t(\mathbf{v}_t)$. We verify in Suppl. that the precision of pseudo-labels for target samples in $\mathcal{B}_t^c$ is high. Thus, 
 the loss $\mathcal{L}_c$ along with $\mathcal{L}_{r1}$ can be viewed as conditioning the classifier $\{g_s, g_t\}$ to deliver a performance close to that of the optimal joint classifier (with the minimal risk $\lambda$).

\textbf{e) Optimization.} We pre-train $\{f_e, f_d\}$ 
to a near-identity function with the losses $l_2(\mathbf{u}_s^c, f_e(\mathbf{u}_s^c))^2$ and $l_2(\mathbf{u}_s^c, f_d(\mathbf{u}_s^c))^2$, where $\mathbf{u}_s \sim \mathcal{P}_s^c ~\forall c \in \mathcal{C}_s$ and $l_2$ is the euclidean distance (similar to an auto-encoder). The total loss employed is $\mathcal{L}_t = \mathcal{L}_a + \mathcal{L}_c + \mathcal{L}_r$, which tightens the bound in Eq.~\ref{upper_bound} as argued above, yielding a superior adaptation guarantee. Instead of directly enforcing $\mathcal{L}_t$ at each iteration, we alternatively optimize each loss using separate Adam~\cite{adam} optimizers in a round robin fashion (\ie we cycle through the losses $\{\mathcal{L}_{a1}, \mathcal{L}_{a2}, \mathcal{L}_c, \mathcal{L}_{r1}, \mathcal{L}_{r2}\}$ and minimize a single loss at each iteration). Since each optimizer minimizes its corresponding loss function independently, the gradients pertaining to each loss are adaptively scaled via the higher order moments~\cite{adam}. This allows us to avoid hyperparameter search for loss scaling. See Suppl. for the training algorithm.

\section{Experiments}
\label{sec:experiments}

We conduct experiments on three datasets. \textbf{Office}~\cite{office} is the most popular benchmark containing 31 classes across 3 domains - Amazon (\textbf{A}), DSLR (\textbf{D}) and Webcam (\textbf{W}). \textbf{VisDA}~\cite{visda} contains 12 classes with 2 domains - Synthetic (\textbf{Sy}) and Real (\textbf{Re}) with a large domain-shift. \textbf{Digits} dataset is composed of MNIST (\textbf{M}), SVHN (\textbf{S}) and USPS (\textbf{U}) domains. See Suppl. for label-set details.

\textbf{a) Evaluation.} We consider two setups for target-private samples - i) one-shot, and, ii) few-shot (5\% labeled). In both cases, we report the mean target accuracy over $\mathcal{C}_t$ (ALL) and $\mathcal{C}_t^{\prime}$ (PRIV), over 5 separate runs (with randomly chosen one-shot and few-shot samples). We compare against prior UDA methods DANN~\cite{ganin2016domain}, OSBP~\cite{Saito_2018_ECCV}, UAN~\cite{UDA_2019_CVPR}, STA~\cite{sta_open_set}, and CI methods E2E~\cite{IncrementalCastro_2018_ECCV}, LETR~\cite{feifei-incremental}, iCaRL~\cite{icarl}, LwF-MC~\cite{lwm}, LwM~\cite{lwm}. To evaluate UDA methods in CIDA, we collect the target samples predicted as \textit{unknown} after adaptation.
We annotate a few of these samples following the few-shot setting, and train a separate target-private classifier (TPC) similar in architecture to $g_t$. At test time, a target sample is first classified into $\mathcal{C}_s \cup \{\textit{unknown}\}$, and if predicted as \textit{unknown}, it is further classified by the target-private classifier. We evaluate the prior arts only in the few-shot setting since they require labeled samples for reliable model upgrade.

\textbf{b) Implementation.} See Suppl. for the architectural details and an overview of the training algorithms for each stage. A learning rate of $0.0001$ is used for the Adam~\cite{adam} optimizers.
For the source-model training, we use 
equal number of source and \textit{negative} samples per batch.
For adaptation, we set $n=20\%$ for confident samples. 
At test time, the prediction for a target sample $\mathbf{x}_t$ is obtained as $\argmax$ over the logits pertaining to $\mathcal{C}_s$ ($g_s \circ f_d \circ f_t (\mathbf{x}_t)|_{c \in \mathcal{C}_s}$) and $\mathcal{C}_t$ ($g_t \circ f_t (\mathbf{x}_t)$).

\subsection{Discussion}
\label{sec:discussion}

\textbf{a) Baseline Comparisons.} To empirically verify the effectiveness of our approach, we implement the following baselines. See Suppl. for illustrations of the architectures. The results are summarized in Table~\ref{tab:baselines}.

i) \textit{{Ours-a}}: To corroborate the need for a target-specific feature space, we remove $\{f_e, f_d\}$, and discard the loss $\mathcal{L}_{r2}$. Here, the $\mathcal{V}$-space is common to both the target and the proxy-source samples. Thus, the \textit{guides} for the shared classes are the fixed class-means ($\boldsymbol{\mu}_s^c$), and the only trainable components are $f_t$ and $g_t$.
In doing so, we force the target classes to acquire the semantics of the source domain which hinders the placement of target-private classes and degrades the target-private accuracy. However, in our approach (\textit{Ours}), trainable \textit{guides} allow the rearrangement of features which effectively minimizes the $d_{\mathcal{H}\Delta\mathcal{H}}$ (in Eq.~\ref{upper_bound}).
    
ii) \textit{{Ours-b}}: To study the regularization of the sampled proxy-source features, we modify our approach by removing $\mathcal{L}_{r}$. We observe a consistent degradation in performance resulting from a lower target-shared accuracy.
    This verifies the role of $\mathcal{L}_{r}$ in mitigating catastrophic forgetting (\ie by minimizing $\epsilon_s$ in Eq.~\ref{upper_bound}).
    
iii) \textit{{Ours-c}}: We modify our approach by removing  $\mathcal{L}_{a2}$ that produces compact target clusters. 
    We find that the target-private accuracy decreases, verifying the need for compact clusters to preserve the semantic granularity across the target classes. Note, \textit{Ours-c} (having trainable \textit{guides} for $\mathcal{C}_s$) outperforms \textit{Ours-a} (having frozen \textit{guides} for $\mathcal{C}_s$), even in the absence of $\mathcal{L}_{a2}$.

iv) \textit{{Ours-d:}} To establish the reliability of the Gaussian Prototypes, we perform CIDA using the source dataset, \ie using the features $f_s(\mathbf{x}_s)$ instead of the sampled proxy-source features. The performance is similar to \textit{Ours}, confirming the efficacy of the Gaussian Prototypes in modelling the source distribution. This is owed to
$\mathcal{L}_{s1}$ that enhances the reliability of the Gaussian approximation.

\begin{table}[t]
\caption{\textbf{Baseline Comparisons.} Results on \textbf{Office}, \textbf{Digits} and \textbf{VisDA} for CIDA using \textbf{one-shot} target-private samples.
}

\centering
\renewcommand{\arraystretch}{1.2}
\setlength{\tabcolsep}{5pt}
\resizebox{1\textwidth}{!}{
\begin{tabular}{|c||c|c||c|c||c|c||c|c||c|c||c|c||c|c|}

\hline
\multirow{3}{*}{\textbf{Method}} & \multicolumn{14}{c|}{\cellcolor{gray!7} 
\textbf{Office} $|\mathcal{C}_s|=20, |\mathcal{C}_t|=31$ } \\

\cline{2-15}
 & \multicolumn{2}{c||}{\textbf{A$\rightarrow$D}} & \multicolumn{2}{c||}{\textbf{A$\rightarrow$W}}  & \multicolumn{2}{c||}{\textbf{D$\rightarrow$A}} & \multicolumn{2}{c||}{\textbf{D$\rightarrow$W}} & \multicolumn{2}{c||}{\textbf{W$\rightarrow$A}} & \multicolumn{2}{c||}{\textbf{W$\rightarrow$D}} & \multicolumn{2}{c|}{\textbf{Avg}} \\

  \cline{2-15}

& ALL & PRIV & ALL & PRIV & ALL & PRIV & ALL & PRIV & ALL & PRIV & ALL & PRIV & ALL & PRIV \\

\hline

\textit{Ours-a} & 69.1 & 66.7 & 58.2 & 55.9 & 60.6 & 58.1 & 70.2 & 68.9 & 59.4 & 57.6 & 80.4 & 80.0 & 66.4 & 64.5 \\
\textit{Ours-b} & 69.5 & 71.9 & 58.2 & 60.4 & 60.9 & 61.1 & 73.3 & 75.6 & 61.1 & 62.3 & 81.7 & 82.8 & 67.5 & 69.0 \\
\textit{Ours-c} & 70.4 & 70.1 & 60.4 & 58.7 & 61.7 & 60.4 & 75.4 & 73.8 & 61.7 & 61.2 & 85.9 & 84.8 & 69.3 & 68.1 \\
\textit{Ours-d} & 73.7 & 73.6 & 64.8 & 64.6 & 64.4 & 63.9 & 80.9 & 80.7 & 63.6 & 61.8 & 90.1 & 89.6 & 72.9 & 72.4 \\
\hline
\textit{Ours} & 73.3 & 73.1 & 63.6 & 62.6 & 64.1 & 64.3 & 80.3 & 79.4 & 63.7 & 62.8 & 89.5 & 88.4 & 72.4 & 71.8 \\

\hline
\hline
\multirow{3}{*}{\textbf{Method}} & \multicolumn{8}{c||}{ \cellcolor{gray!7} 
\textbf{Digits} ($|\mathcal{C}_s|=5, |\mathcal{C}_t|=10$)} & \multicolumn{6}{c|}{ \cellcolor{gray!7} \textbf{VisDA} ($|\mathcal{C}_s|=6, |\mathcal{C}_t|=12$)}\\

\cline{2-15}
  & \multicolumn{2}{c||}{\textbf{S$\rightarrow$M}} & \multicolumn{2}{c||}{\textbf{M$\rightarrow$U}} & \multicolumn{2}{c||}{\textbf{U$\rightarrow$M}} & \multicolumn{2}{c||}{\textbf{Avg}} & \multicolumn{2}{c||}{\textbf{Sy$\rightarrow$Re}} & \multicolumn{2}{c||}{\textbf{Re$\rightarrow$Sy}} & \multicolumn{2}{c|}{\textbf{Avg}} \\
\cline{2-15}

& ALL & PRIV & ALL & PRIV & ALL & PRIV & ALL & PRIV & ALL & PRIV & ALL & PRIV & ALL & PRIV \\

\cline{1-15}

\textit{Ours-a} & 41.2 & 38.7 & 64.9 & 63.5 & 63.1 & 62.6 & 56.4 & 54.9 & 52.3 & 51.4 & 50.9 & 49.6 & 51.6 & 50.5 \\
\textit{Ours-b} & 42.4 & 42.9 & 66.1 & 67.2 & 63.9 & 64.7 & 57.5 & 58.3 & 53.1 & 53.6 & 51.1 & 51.4 & 52.1 & 52.5 \\
\textit{Ours-c} & 44.5 & 43.8 & 69.4 & 69.3 & 65.3 & 64.5 & 59.7 & 59.2 & 54.3 & 54.0 & 52.3 & 51.9 & 53.3 & 52.9 \\
\textit{Ours-d} & 46.5 & 45.3 & 72.7 & 72.2 & 69.4 & 68.8 & 62.9 & 62.1 & 56.6 & 56.3 & 55.8 & 55.4 & 56.2 & 55.8 \\
\hline
\textit{Ours} & 46.4 & 45.7 & 72.5 & 71.6 & 69.4 & 68.6 & 62.8 & 61.9 & 56.4 & 56.3 & 55.8 & 55.7 & 56.1 & 56.0 \\

\hline
\end{tabular}}

\label{tab:baselines}
\end{table}

\begin{table}[t]
\caption{
\noindent \textbf{Comparison against prior arts.} Results on \textbf{Office} ($|\mathcal{C}_s|=10, |\mathcal{C}_t|=20$) for CIDA. Unsup. denotes the method is unsupervised (on target). SF denotes model upgrade is \textit{source-free}. Note, non-\textit{source-free} methods access labeled source data.
Results are grouped based on access to i) few-shot and ii) one-shot target-private samples. 
}

\label{tab:incrementalOneShotComp}
\setlength\tabcolsep{4pt}
\renewcommand{\arraystretch}{1.1}

\begin{center}
\begin{small}
\begin{sc}

\resizebox{1\textwidth}{!}{
\begin{tabular}{|c|c|c||c|c||c|c||c|c||c|c||c|c||c|c||c|c|}

\hline
\multirow{2}{*}{\textbf{Method}} & \multirow{2}{*}{\textbf{SF}} & 
\multirow{2}{*}{\textbf{Unsup.}} & \multicolumn{2}{c||}{\textbf{A$\rightarrow$D}} & \multicolumn{2}{c||}{\textbf{A$\rightarrow$W}} & \multicolumn{2}{c||}{\textbf{D$\rightarrow$A}} & \multicolumn{2}{c||}{\textbf{D$\rightarrow$W}} & \multicolumn{2}{c||}{\textbf{W$\rightarrow$A}} & \multicolumn{2}{c||}{\textbf{W$\rightarrow$D}} & \multicolumn{2}{c|}{\textbf{Avg}} \\
\cline{4-17} 

& & & all & priv & all & priv & all & priv & all & priv & all & priv & all & priv & all & priv \\
\hline

\rowcolor{gray!7}
\multicolumn{17}{|c|}{\textbf{Using few-shot target-private samples (5\% labeled)}}\\

\hline

DANN+TPC & \xmark & \cmark & 54.3 & 21.4 & 52.4 & 16.7 & 48.2 & 19.8 & 61.4 & 24.3 & 57.9 & 21.1 & 56.5 & 38.0 & 55.1 & 23.6\\
OSBP+TPC & \xmark & \cmark & 51.6 & 13.9 & 49.2 & 9.5 & 58.8 & 14.3 & 55.5 & 18.4 & 49.1 & 13.6 & 64.0 & 29.1 & 54.7 & 14.5\\
STA+TPC & \xmark & \cmark & 56.6 & 17.7 & 51.2 & 10.2 & 54.8 & 16.5 & 59.6 & 21.8 & 54.7 & 15.9 & 67.4 & 35.4 & 57.4 & 19.6\\
UAN+TPC & \xmark & \cmark & 56.2 & 24.4 & 54.8 & 21.2 & 57.3 & 24.7 & 62.6 & 29.5 & 59.2 & 28.9 & 68.4 & 42.8 & 59.8 & 28.6\\
\hline
iCaRL & \xmark & \xmark & 63.6 & 63.2 & 54.3 & 53.8 & 56.9 & 56.1 & 65.4 & 65.2 & 57.5 & 56.8 & 76.8 & 77.5 & 62.4 & 62.1\\
E2E & \xmark & \xmark & 64.2 & 61.9 & 55.6 & 53.2 & 58.8 & 58.4 & 66.3 & 66.5 & 57.9 & 56.6 & 76.5 & 73.2 & 63.2 & 60.8\\
LETR & \xmark & \cmark & 71.3 & 68.5 & 58.4 & 57.6 & 58.2 & 58.4 & 70.3 & 69.8 & 62.0 & 60.7 & 84.2 & 82.9 & 67.4 & 66.3\\
LwM & 
\cmark & \xmark & 66.5 & 66.2 & 56.3 & 55.9 & 57.6 & 56.8 & 68.4 & 68.3 & 59.8 & 59.4 & 78.4 & 78.1 & 64.5 & 61.9\\
LwF-MC & 
\cmark & \xmark & 64.3 & 63.8 & 55.6 & 55.1 & 55.5 & 55.7 & 67.6 & 67.7 & 59.4 & 59.0 & 77.3 & 76.9 & 63.3 & 63.0\\
\hline

\textit{Ours*} & \cmark & \cmark & \textbf{78.8} & \textbf{74.3} & \textbf{70.1} & \textbf{69.8} & \textbf{66.9} & \textbf{67.1} & \textbf{85.0} & \textbf{84.6} & \textbf{67.2} & \textbf{65.3} & \textbf{90.4} & \textbf{90.8} & \textbf{76.4} & \textbf{75.3} \\

\hline

\rowcolor{gray!7}
\multicolumn{17}{|c|}{\textbf{Using one-shot target-private samples}} \\

\hline

\textit{Ours-a} & \cmark & \cmark & 67.4 & 64.1 & 56.2 & 53.4 & 60.1 & 57.8 & 69.2 & 68.3 & 57.1 & 55.6 & 77.9 & 76.5 & 65.0 & 62.3 \\
\textit{Ours-b} & \cmark & \cmark & 68.4 & 70.2 & 57.5 & 59.6 & 60.6 & 60.8 & 70.9 & 72.4 & 58.4 & 58.7 & 79.8 & 80.2 & 65.9 & 67.0 \\
\textit{Ours-c} & \cmark & \cmark & 70.0 & 69.3 & 59.5 & 57.4 & 61.5 & 60.2 & 73.1 & 71.4 & 61.8 & 60.1 & 82.3 & 81.1 & 68.0 & 66.6 \\
\hline
\textit{Ours} & \cmark & \cmark & \textbf{72.2} & \textbf{72.6} & \textbf{62.1} & \textbf{62.0} & \textbf{62.6} & \textbf{61.8} & \textbf{78.5} & \textbf{78.7} & \textbf{62.1} & \textbf{62.4} & \textbf{87.8} & \textbf{87.6} & \textbf{70.7} & \textbf{70.8} \\

\hline

\end{tabular}
}

\end{sc}
\end{small}
\end{center}
\end{table}

\noindent
\textbf{b) Comparison against prior arts.} We compare against prior UDA and CI approaches in Table~\ref{tab:incrementalOneShotComp}. 
Further, we run a variation of our approach with few-shot (5\% labeled) target-private samples (\textit{Ours*}), where the \textit{guides} for $\mathcal{C}_t^{\prime}$ are obtained as the class-wise mean features of the few-shot samples. 

UDA methods exploit unlabeled target samples but require access to labeled source samples during adaptation. They achieve a low target-private ({PRIV}) accuracy owing to the loss of semantic granularity. This effect is evident in open-set methods, where target-private samples are forced to be clustered into a single \textit{unknown} class. However in DANN and UAN, such a criterion is not enforced, instead a target sample is detected as \textit{unknown} using confidence thresholding. Thus, DANN and UAN achieve a higher {PRIV} accuracy than STA and OSBP.

The performance of most CI methods in CIDA is limited due to the inability to address domain-shift.
LETR, E2E and iCaRL require labeled samples from both the domains during the model upgrade.
E2E exploits these labeled samples to re-train the source-trained model with all classes ($\mathcal{C}_t$). However, the need to generalize across two domains degrades the performance on the target domain where target-shared samples are unlabeled. 
In contrast, LwM and LwF-MC learn a separate target model, by employing a distillation loss using the target samples.
However, distillation is not suitable under a domain-shift since the source model is biased towards the source domain characteristics that cannot be generalized for the target domain. In LETR, the global domain statistics across the two domains are aligned. However, such a global alignment is prone to the negative influence of target-private samples which limits its performance.

Our method addresses these limitations and outperforms both UDA and CI methods. The foresighted source-model training suppresses domain and category bias by addressing the overconfidence issue. Then, a gradual rearrangement of features in a target-specific semantic space allows the learning of target-private classes while preserving the semantic granularity. Furthermore, the regularization from the proxy-source samples mitigates catastrophic forgetting. Thus our approach achieves a more stable performance in CIDA, even in the challenging \textit{source-free} scenario. See Suppl. for a discussion from the theoretical perspective.

\noindent 
\textbf{c) Effect of class separability objective.} 
We run an ablation on the \textbf{A$\rightarrow$D} task (Table~\ref{tab:incrementalOneShotComp}) without enforcing $\mathcal{L}_{s1}$ during the source-model training. The accuracy post adaptation is 68.6\% ({PRIV} = 70.4\%) as compared to $72.2\%$ ({PRIV} = 72.6\%) in \textit{Ours}. This suggests that the \textit{class separability} objective (enforcing the Cluster Assumption) helps in generalization to the target domain.

\noindent
\textbf{d) Effect of negative training.} On the \textbf{A$\rightarrow$D} task (Table~\ref{tab:incrementalOneShotComp}), a source-model trained with \textit{negative} samples ($\mathcal{D}_s \cup \mathcal{D}_n$) achieves a source accuracy of 96.7\%, while that trained without \textit{negative} samples yields 96.9\%.
Thus, there is no significant drop on the source performance due to negative training. However, this aids in generalizing the model to novel target classes. Specifically, a source-model trained with \textit{negative} samples (\textit{Ours}) yields 72.2\% ({PRIV} = 72.6\%) after adaptation, while that without \textit{negative} training achieves 67.4\% ({PRIV} = 62.3\%) after adaptation. The performance gain in \textit{Ours} is attributed to the mitigation of the overconfidence issue thereby reliably classifying target-private samples. 

\begin{figure}[t!]
    \centering
    \includegraphics[width=1\linewidth]{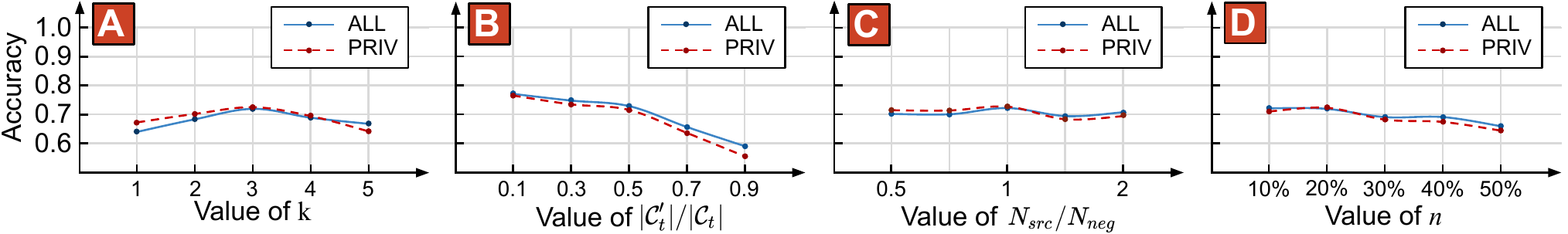}
    \caption{
    {\textbf{Sensitivity for A$\rightarrow$D task (Office).} \textbf{A)} Confidence interval $k$-$\sigma$ for \textit{negative} sampling. \textbf{B)} Fraction of target-private classes $|\mathcal{C}_t^{\prime}|/|\mathcal{C}_t|$ during CIDA. \textbf{C)} Batch size ratio of source ($N_{src}$) and \textit{negative} ($N_{neg}$) samples during source-training. \textbf{D)} Percentage of confident target samples for $\mathcal{B}_t^c$ during CIDA. Note the scale of the axes. \vspace{-1mm}
    }                                                           
    }
    \label{fig:sensitivity}
\end{figure}

\noindent
\textbf{e) Sensitivity to hyperparameters.} In Fig.~\ref{fig:sensitivity}, we plot the target accuracy post adaptation for various hyperparameter values for the task \textbf{A$\rightarrow$D}. Empirically, we found that a 3-$\sigma$ confidence interval for \textit{negative} sampling was most effective in capturing the source distribution (Fig.~\ref{fig:sensitivity}{\color{BUred}A}). We choose an equal number of source ($N_{src}$) and \textit{negative} ($N_{neg}$) samples in a batch during source training to avoid the bias caused by imbalanced data. Fig.~\ref{fig:sensitivity}{\color{BUred}C} shows the sensitivity to the batch size ratio $N_{src} / N_{neg}$. Further, the hyperparameter $n$ is marginally stable around $n=20\%$ (Fig.~\ref{fig:sensitivity}{\color{BUred}D}) which was used across all experiments. Finally, the trend in Fig.~\ref{fig:sensitivity}{\color{BUred}B} is a result of the challenging one-shot setting.

\noindent
\textbf{f) Two-step model upgrade.}
We extend our approach to perform two-step model upgrade under CIDA on \textbf{Office} (See Suppl. for details). First a source model ($\{f_s, g_s\}$) is trained on the 10 classes of Amazon (\textbf{A}) which is upgraded to the 20 classes of DSLR (\textbf{D}) thereby learning $\{f_t, g_t, f_e, f_d\}$.
We upgrade this DSLR-specific model to the Webcam (\textbf{W}) domain, having 20 classes shared with (\textbf{A+D}), and 11 new classes. This is done by learning feature extractor $f_{t_2}$, classifier $g_{t_2}$, and domain projection networks $\{f_{e_2}, f_{d_2}\}$ learned between the latent spaces of $f_t$ and $f_{t_2}$. We observe an accuracy of 79.9\% on \textbf{W}, which is close to that obtained by directly adapting from 20 classes of DSLR to 31 classes in Webcam (80.3\%, Table \ref{tab:baselines}). This corroborates the practical applicability of our approach to multi-step model upgrades. See Suppl. for a detailed discussion.

\vspace{-2mm}
\section{Conclusion}

We proposed a novel Domain Adaptation paradigm (CIDA) addressing class-incremental learning in the presence of domain-shift.
We studied the limitations of prior approaches in the CIDA paradigm and proposed a two-stage approach to address CIDA.
We presented a foresighted source-model training that facilitates a \textit{source-free} model upgrade.
Further, we demonstrated the efficacy of a target-specific semantic space, learned using trainable \textit{guides}, that preserves the semantic granularity across the target classes. Finally, our approach shows promising results on multi-step model upgrades. 
As a future work, the framework can be extended to a scenario where a series of domain-shifts and task-shifts are observed.

\noindent\textbf{Acknowledgement.} This work is supported by a Wipro PhD Fellowship and a grant from Uchhatar Avishkar Yojana (UAY, IISC\_010), MHRD, Govt. of India.

\clearpage



\section*{Supplementary material}

# Supplementary Material

## Class-Incremental Domain Adaptation

| Reference in paper | Contents | Section in Suppl. |
|---|---|---|
| §2 | Experimental details of STA | 1 |
| §3 | Notation Table | 2 |
| §3 | Algorithms | 3 |
| §3.1b | Generation of *negative* samples ($\mathbf{u}_n$) | 3.1 |
| §3.1c | Foresighted source-model training | 3.2 |
| §3.2c | Selection of confident target samples ($\mathcal{B}_t^c$) | 3.3 |
| §3.2e | Class-Incremental Domain Adaptation | 3.4 |
| §3.2d | Pseudo-labeling precision | 4 |
| §4.1 | Dataset (label-set) details | 5 |
| §4 | Architecture details | 6 |
| §4.1b | **Office** and **Visda** datasets | 6.1 |
| §4.1b | **Digits** dataset | 6.2 |
| §4.2a | Illustrations for baselines | 6.3 |
| §4.2b | Theoretical Insights | 7 |
| §4.2f | Discussion on multi-step model upgrade | 8 |

## 1 Experimental details of STA

The label-sets used for the experiment are as mentioned in Sec. 5 and the architecture is given in [7]. After the adaptation, we predict labels for all target samples and collect the samples predicted as *unknown*. We then label 5% of these samples (ensuring equal number of samples in each class) to obtain few-shot target-private samples. Note that the few-shot target-private samples do not contain the mis-classified target-shared samples. Finally, we learn a target-private classifier (TPC) that is similar to $g_t$ (Table 2).

We evaluate the method over 5 separate runs (with different sets of few-shot samples). First, we classify all target samples into $\mathcal{C}_s \cup \{unknown\}$. Then, samples predicted as *unknown* are further classified into $\mathcal{C}_t'$ using the TPC, thereby assigning a unique target-private label to each sample predicted as *unknown*. With target samples classified into $|\mathcal{C}_t|$ classes, we obtain the accuracy on target-shared samples (95.9 ± 0.3%) and target-private samples (17.7 ± 3.5%).

This evaluation scheme is followed for all UDA methods. Particularly, in DANN [3] and UAN [13], we obtain the *unknown* label prediction using confidence thresholding. In STA [7] and OSBP [12], the classifier is trained with an additional *unknown* class that is used to detect *unknown* samples.

## 2   Notation Table

In Table 1, we provide an exhaustive list of symbols used in the paper and the corresponding description.

**Table 1.** Notations used in the paper

| | Symbol | Description |
|---|---|---|
| Space | $\mathcal{X}$ | Input space |
| | $\mathcal{Y}$ | Label space |
| | $\mathcal{U}$ | Output space of $f_s$ |
| | $\mathcal{V}$ | Output space of $f_t$ |
| Distr. | $p$ | Source distribution on $\mathcal{X} \times \mathcal{Y}$ |
| | $q$ | Target distribution on $\mathcal{X} \times \mathcal{Y}$ |
| | $q_\mathcal{X}$ | Target marginal input distribution |
| | $r$ | *Negative* data distribution in $\mathcal{U}$-space |
| Network | $f_s$ | Feature extractor of the source-model |
| | $g_s$ | Classifier of the source-model |
| | $f_t$ | Target-specific feature extractor |
| | $g_t$ | Classifier accounting for $\mathcal{C}'_t$ |
| | $f_e, f_d$ | Domain projection networks |
| Sets | $\mathcal{D}_s$ | Set of labeled source samples |
| | $\mathcal{D}_t$ | Set of unlabeled target samples |
| | $\mathcal{D}_n$ | Set of labeled *Negative* samples |
| | $\mathcal{C}_s, \mathcal{C}_t, \mathcal{C}'_t$ | Set of shared classes, target classes and target-private classes |
| | $\mathcal{B}_t^c$ | Set of confident target samples for each class $c \in \mathcal{C}_t$ |
| Samples | $\mathbf{x}_s, \mathbf{x}_t$ | Source and Target domain input samples (image) |
| | $y_s, y_t$ | Source and Target domain labels |
| | $\{(\tilde{\mathbf{x}}_t^c, \tilde{y}_t^c)\}$ | One-shot target-private samples $\forall c \in \mathcal{C}'_t$ |
| | $\hat{y}(\cdot)$ | Concatenated output pertaining to $\mathcal{C}_s$ ($g_s(\cdot)|_{c \in \mathcal{C}_s}$) and $\mathcal{C}_t$ ($g_t$) |
| | $\mathbf{u}_s$ | Source / proxy-source sample feature in the $\mathcal{U}$-space |
| | $\mathbf{u}_n$ | *Negative* sample feature in the $\mathcal{U}$-space |
| | $\mathbf{v}_t$ | Target sample feature in the $\mathcal{V}$-space |
| | $\mathbf{v}_g^c$ | *Guide* for class $c \in \mathcal{C}_t$ in the $\mathcal{V}$-space |
| Prototypes | $\mathcal{P}_s^c, \mathcal{P}_t^c$ | Source and Target domain class-specific Gaussian Prototypes |
| | $\mathcal{P}_s, \mathcal{P}_t$ | Source and Target domain Global Gaussian Prototypes |
| | $\boldsymbol{\mu}_s^c, \boldsymbol{\Sigma}_s^c$ | Class-specific mean and covariance of source features $f_s(\mathbf{x}_s)$ |
| | $\boldsymbol{\mu}_s, \boldsymbol{\Sigma}_s$ | Mean and covariance of source features $f_s(\mathbf{x}_s)$ for all $\mathbf{x}_s$ in $\mathcal{D}_s$ |
| | $\boldsymbol{\mu}_t^c, \boldsymbol{\Sigma}_t^c$ | Class-specific mean and covariance of target features $f_t(\mathbf{x}_t)$ |
| | $\boldsymbol{\mu}_t, \boldsymbol{\Sigma}_t$ | Mean and covariance of source features $f_t(\mathbf{x}_t)$ for all $\mathbf{x}_t$ in $\mathcal{D}_t$ |
| Misc. | $\mathbb{E}$ | Expectation |
| | $l_{ce}(\cdot, \cdot)$ | Standard cross-entropy function |
| | $l_2(\cdot, \cdot)$ | Euclidean distance function |
| | $d, k$ | Distance to the nearest *guide* and pseudo-label of a target sample |
| | $n$ | Percentage of confident target samples during CIDA |
| | $N_{src}, N_{neg}$ | Number of source and *negative* samples in a batch |

## 3  Algorithms

Here we provide the algorithms corresponding to each stage of our approach, along with a description of the major steps involved.

### 3.1  Generation of *negative* samples ($\mathbf{u}_n$)

During the source-model training, we generate *negative* samples at each iteration as outlined in Algo. 1. Specifically, *negative* samples are those samples obtained from the global Gaussian Prototype $\mathcal{P}_s = \mathcal{N}(\boldsymbol{\mu}_s, \boldsymbol{\Sigma}_s)$ that fall outside the 3-$\sigma$ confidence threshold of each class-specific Gaussian Prototype $\mathcal{P}_s^c = \mathcal{N}(\boldsymbol{\mu}_s^c, \boldsymbol{\Sigma}_s^c)$.

This 3-$\sigma$ threshold is calculated in lines 3-6 of Algo. 1, where we obtain the eigen decomposition of the covariance matrix $\boldsymbol{\Sigma}_s^c$ and select the largest eigen value $\lambda_s^c$ and the corresponding eigen vector $\boldsymbol{\nu}_s^c$. Then, we define the 3-$\sigma$ threshold for class $c$ as the likelihood of the sample $\boldsymbol{\mu}_s^c + 3 \cdot \sqrt{\lambda_s^c} \cdot \boldsymbol{\nu}_s^c$.

To generate *negative* samples ($\mathbf{u}_n$), we obtain samples from the global Gaussian Prototype $\mathcal{P}_s$ (line 8) and reject those samples that are within the 3-$\sigma$ confidence interval of any class. In other words, we keep those samples which do not fall within the 3-$\sigma$ interval of any class (lines 10-12).

### 3.2  Foresighted source-model training

The source-model training is outlined in Algo. 2. We pre-train the network for 1 epoch (lines 2-5) using the vanilla cross-entropy loss (to provide a fair initialization for the Gaussian Prototypes). Here, line 4 denotes the parameter update step performed by gradient descent using the Adam optimizer. The term inside the *Adam* operator denotes the (negative) gradient of the mean loss for the mini-batch of the source samples.

After this initialization, we calculate the Gaussian Prototypes, both class-specific (lines 6-11) and global (lines 12-14). Note that, in the paper we mention that an alternating minimization scheme is used for training the network. This is implemented by maintaining a list of corresponding losses, optimizers and model parameters. Particularly, use a separate *Adam* optimizer (line 16) for each loss (line 15) that trains only the parameters specific to the loss (line 17). Further, we maintain a counter *iter* (line 18) that is used to determine the loss that is to be optimized at each iteration ($cur = iter$ mod 2).

During training, we obtain a mini-batch with an an equal amount of source ($N_{src}$) and *negative* samples ($N_{neg}$), i.e. $N_{src} = N_{neg}$ (lines 22-29). Line 30 denotes the optimization process, where $cur$ is used to index the list of parameters $\boldsymbol{\Theta}$, optimizers **Opt** and losses **Loss**. Note, here the update step denotes the update of the parameters $\boldsymbol{\Theta}[cur]$ specific to the loss **Loss**$[cur]$ being optimized by the optimizer **Opt**$[cur]$ (similar to that in line 4). Finally, at the end of each epoch, we recalculate the Gaussian Prototypes which are then used for the next epoch.

### 3.3  Selection of confident target samples $\mathcal{B}_t^c$

The process of confident target sample ($\mathcal{B}_t^c$) selection is shown in Algo. 3. Specifically, for each *guide* $\mathbf{v}_g^c$, we obtain the nearest $n$-percent pseudo-labeled target samples. Lines 3-7 denote the pseudo-labeling process, while the selection of nearest $n$-percent target samples is shown in lines 8-12.

### 3.4  Class-Incremental Domain Adaptation

We outline the algorithm for performing CIDA on the target domain in Algo. 4. The target feature extractor $f_t$ is initialized from $f_s$ and the domain projection networks $f_e, f_d$ are initialized with near-identity mappings (to effectively "initalize $\mathcal{V}$-space with the $\mathcal{U}$-space"). This is depicted in lines 2-7. Particularly, we use separate Adam optimizers ($Adam_{f_e}$ for training $f_e$, and $Adam_{f_d}$ for training $f_d$) and train the networks with the reconstruction losses (similar to a vanilla auto-encoder). Note that, lines 5-6 denote the parameter update step performed by gradient descent using the Adam optimizer specific to each loss. Here, the term inside the *Adam* operator is the (negative) gradient of the mean loss for the mini-batch of the proxy-source samples.

After the initialization, we begin the model upgrade. Note that, as mentioned in the paper, we use separate Adam optimizers for each loss and follow a round-robin alternating minimization scheme by optimizing a single loss using its specific optimizer. This allows each optimizer to adaptively scale the gradients of the corresponding loss, thus, avoiding the need to have separate hyperparameters for loss weighting. To this end, we define lists of corresponding losses (**Loss**), optimizers (**Opt**) and model parameters ($\boldsymbol{\Theta}$) (lines 8-10 in Algo. 4). Further, *iter* (line 11) is a counter that is used to obtain the index of the loss to be optimized at each iteration ($cur = iter \mod 5$).

At each iteration, we obtain the *guides* (line 15) for each class. For the losses $\mathcal{L}_{r1}, \mathcal{L}_{r2}$, we use the proxy-source samples (lines 16-18). The adaptation loss $\mathcal{L}_{a1}$ is applied on the pseudo-labeled target samples (lines 19-22), while the losses $\mathcal{L}_{a2}$ and $\mathcal{L}_c$ are applied on the confident target samples (23-25). For convenience, we depict the round-robin alternating loss optimization in line 26 where for the iteration *iter*, the index of the loss to be optimized is $cur = iter \mod 5$. Thus, the corresponding loss is **Loss**[*cur*], which is optimized using the optimizer **Opt**[*cur*] thereby training the parameters $\boldsymbol{\Theta}$[*cur*]. Note that the implementation of the parameter update step (line 26) is similar to that in lines 5-6, where the actual loss optimized is the mean loss calculated over the samples in the mini-batch under consideration. Finally, at the end of each epoch, we recalculate the Gaussian Prototypes $\mathcal{P}_t^c$ (line 29) and obtain confident target samples $\mathcal{B}_t^c$.

## 4  Pseudo-labeling Precision

We analyze the precision of the pseudo-labels obtained for the top-$n$ percent confident target samples $\mathcal{B}_t^c$ for each class $c \in \mathcal{C}_t$. The precision of the target pseudo-labels for the task **A→D** on **Office** dataset is shown in Fig. 1 (in line with the setting in Table 3 in the paper). The overall precision is computed as the class-averaged precision of the pseudo-labels. Specifically, Fig. 1A shows the precision for samples predicted with a pseudo-label $k \in \mathcal{C}_t$ (ALL) and Fig. 1 shows the precision for samples predicted with a pseudo-label $k \in \mathcal{C}_t'$ (PRIV).

We observe that the pseudo-labeling precision for $n = 20\%$ target samples is high. This validates the efficacy of the pseudo-labeling process. Further, we also show the precision-curve obtained as the training proceeds. We find that the area

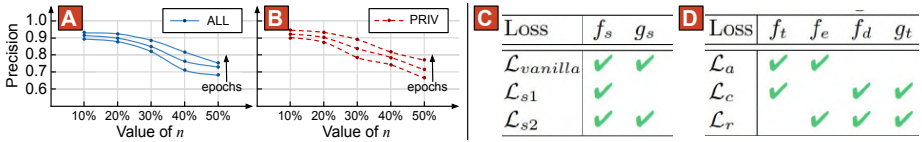

**Fig. 1.** **A)** Pseudo-labeling precision on **A**→**D** for ALL classes ($\mathcal{C}_t$) during adaptation. **B)** Pseudo-labeling precision on **A**→**D** for PRIV classes ($\mathcal{C}'_t$) during adaptation. **C)** Losses imposed during Foresighted source-model training. **D)** Losses imposed during CIDA on the target domain. ✔ indicates the modules trained by each loss.

under the curve increases as the training proceeds (increasing number of epochs is indicated by the arrow ↑). This demonstrates the reliability of pseudo-labels during adaptation. We find this trend in all our experiments, *i.e.* samples that are nearest to the *guides* have the highest pseudo-labeling precision. This validates our intuition for obtaining the set of confident samples $\mathcal{B}^c_t$.

## 5  Dataset details

The **Office** [11] dataset consists of 31 classes and has 3 domains: Amazon (A), DSLR (D) and Webcam(W). In our experiments, we use the first 10 classes in alphabetical order as the shared classes $\mathcal{C}_s$ and those ranked 21-31 alphabetically as target-private classes $\mathcal{C}'_t$. Since the DSLR and Webcam domains contain a relatively small number of images, we perform image augmentations as shown in Fig. 2 for each domain. The **Digits** dataset contains three domains MNIST (M) [6], SVHN (S) [8], USPS (U) [5]. We use the classes from 0 through 4 as the shared categories $\mathcal{C}_s$ whereas the classes from 5 through 9 are used as target-private categories $\mathcal{C}'_t$. The **VisDA** [9] dataset consists of 12 classes in total with Synthetic (Sy) and Real (Re) domains. The source domain consists of synthetic images that were created by rendering CAD models while the target domain consists of real-world images. Here, we set $\mathcal{C}_s =$ {bicycle, bus, car, motorcycle, train, truck}, and set $\mathcal{C}'_t =$ {aeroplane, horse, knife, person, plant, skateboard}.

## 6  Architecture details

In this section, we provide the details of the architectures used for **Office** [11], **VisDA** [9] and **Digits** datasets, along with the architectures of the baseline models presented in the paper (§4.2a). Additionally, in Fig. 1C-D, we show the modules that are trained with each loss imposed during the foresighted source-model training and during CIDA on the target domain.

### 6.1  Office and VisDA datasets

For training on the **Office** and **VisDA** datasets, we use ResNet-50 [4], pretrained on the **ImageNet** [10] dataset, as the CNN backbone for the feature extractor $f_s$. The rest of the network architecture is given in Table 2A.

### 6.2   Digits dataset

We use a modified version of LeNet [5] architecture (Table 2B) for training on **Digits** dataset, similar to that used in [12]. We use RGB images for training on each dataset (for MNIST and USPS, we repeat the pixel values across channels).

### 6.3   Illustrations for baselines

We illustrate the architectures (and gradient backpropagation pathways) for each baseline model in Fig. 3. In *Ours-a*, the clusters of the proxy-source samples in the $\mathcal{V}$-space are frozen (blue clusters in Fig. 3A) and are the same as that obtained in the $\mathcal{U}$-space at the end of the source-model training. In *Ours-b*, the loss $\mathcal{L}_r$ is not imposed as shown in Fig. 3B. In *Ours-c*, the loss $\mathcal{L}_{a2}$ is not enforced, as shown in Fig. 3C. In *Ours-d*, the source features $f_s(\mathbf{x}_s)$ are used in place of the Gaussian Prototypes as shown in Fig. 3D.

## 7   Theoretical Insights

In the paper, we argue that the adaptation process tightens the upper bound of target risk of the model, yielding a superior adaptation guarantee. In this section, we provide a formal discussion for this observation and demonstrate that our approach conforms with the theory proposed in [1]. In our approach, since we aim to learn a target-specific model to perform the model upgrade, we are interested in the dynamics of the target-specific latent space $\mathcal{V}$. Therefore, we begin by analyzing the properties of the distributions in the $\mathcal{V}$-space.

### 7.1   Definitions

Let $q_\mathcal{V}$ denote the marginal distribution of target features in the $\mathcal{V}$-space. Subsequently, we denote the marginal target-shared and target-private distributions by $q_\mathcal{V}^s$ and $q_\mathcal{V}^p$ respectively. Further, let the marginal distribution of the high-source-density region (effectively modeled by proxy-source samples) be $p_\mathcal{V}$, and the marginal distribution corresponding to the low-source-density region (effectively modeled by *negative* samples) be $r_\mathcal{V}$.

In our formulation, for a given sample, the label is predicted by taking the arg max over the logits corresponding to $\mathcal{C}_s$ and $\mathcal{C}_t$. In other words, for a sample $\mathbf{v}$ in the $\mathcal{V}$-space, we can define the label predictor (classifier) as follows,

$$g(\mathbf{v}) = \arg\max_{c \in \mathcal{C}_t} \ \ g_s \circ f_d(\mathbf{v})|_{c' \in \mathcal{C}_s} \ ||\ g_t(\mathbf{v}) \tag{1}$$

where $g \in \mathcal{H}$ is an instance of predictors pertaining to the hypothesis space $\mathcal{H}$ and $||$ denotes concatenation. With this setup, we derive an upper bound on the target-domain risk of the predictor $g$ and show that the bound is tightened by the adaptation process. To this end, we define two useful tools - the disagreement among predictors, and, the triangle inequality for classification error.

**Definition 1 (Disagreement Measure).** *The disagreement between two predictors $g_1$ and $g_2$ under a distribution $d$ is given as,*

$$\epsilon_d(g_1, g_2) = \Pr_{\mathbf{v} \sim d}\Big(g_1(\mathbf{v}) \neq g_2(\mathbf{v})\Big) \tag{2}$$

Eq. 2 is a multi-class variation of the classifier disagreement proposed in [1]. We now extend Def. 1 to define the risk of a predictor.

**Definition 2 (Risk Measure).** *Let the ground-truth labeling function be $h$. The risk of a predictor $g$ under a distribution $d$ is defined as the probability with which the classifier $g$ disagrees with the ground-truth labeling function $h$,*

$$\epsilon_d(g, h) = \Pr_{\mathbf{v} \sim d}\Big(g(\mathbf{v}) \neq h(\mathbf{v})\Big) \tag{3}$$

For convenience, we shall use the short-hand notation to denote the risk of a predictor $g$ as $\epsilon_d(g)$, i.e. , $\epsilon_d(g) = \epsilon_d(g, h)$. Further, we use the disagreement measure to define the $\mathcal{H}\Delta\mathcal{H}$-distance [1] between two distributions $d_1$ and $d_2$.

**Definition 3 ($\mathcal{H}\Delta\mathcal{H}$-distance).** *Given a hypothesis space $\mathcal{H}$ and two distributions $d_1$ and $d_2$, the $\mathcal{H}\Delta\mathcal{H}$-distance is given by*

$$l_{\mathcal{H}\Delta\mathcal{H}}(d_1, d_2) = 2 \sup_{g_1, g_2 \in \mathcal{H}} |\epsilon_{d_1}(g_1, g_2) - \epsilon_{d_2}(g_1, g_2)| \tag{4}$$

Here, $l_{\mathcal{H}\Delta\mathcal{H}}$ is a measure of the distribution shift between the distributions $d_1$ and $d_2$. In other words, $l_{\mathcal{H}\Delta\mathcal{H}}$ grows as the discrepancy between the two distributions (domain-shift) grows. Note, in this section we use the notation $l_{\mathcal{H}\Delta\mathcal{H}}$ in place of $d_{\mathcal{H}\Delta\mathcal{H}}$ as done in [1], since here, $d$ denotes a generic distribution. Now we describe the second tool, the triangle inequality for classification error [2].

**Lemma 1 (Triangle Inequality).** *Given three predictors $g_1, g_2, g_3 \in \mathcal{H}$, the triangle inequality for the disagreement measure is stated as*

$$\epsilon_d(g_1, g_3) \leq \epsilon_d(g_1, g_2) + \epsilon_d(g_2, g_3) \tag{5}$$

Finally, let us describe the performance of an ideal predictor. We define a predictor as ideal if it achieves the minimal combined risk over all distributions under consideration.

**Definition 4 (Ideal Predictor).** *Given a hypothesis space $\mathcal{H}$ and a set of $n$ distributions $\{d_1, d_2, ...d_n\}$, an ideal predictor $g* \in \mathcal{H}$ achieves the minimal combined risk on the given set of distributions. In other words,*

$$g* = \arg\min_{g \in \mathcal{H}} \sum_{i=1}^{n} \epsilon_{d_i}(g) \tag{6}$$

Particularly, we consider the following distributions in the $\mathcal{V}$-space: the source distribution $p_\mathcal{V}$, the *negative* distribution $r_\mathcal{V}$, the target-shared distribution $q_\mathcal{V}^s$ and the target-private distribution $q_\mathcal{V}^p$. Thus, the ideal predictor $g*$ is given as,

$$g* = \arg\min_{g \in \mathcal{H}} \epsilon_{p_\mathcal{V}}(g) + \epsilon_{r_\mathcal{V}}(g) + \epsilon_{q^s_\mathcal{V}}(g) + \epsilon_{q^p_\mathcal{V}}(g) \qquad (7)$$

Note that, given a hypothesis space $\mathcal{H}$, the combined risk of the ideal predictor $g*$ is a constant (*i.e.* $\lambda = \epsilon_{p_\mathcal{V}}(g*) + \epsilon_{r_\mathcal{V}}(g*) + \epsilon_{q^s_\mathcal{V}}(g*) + \epsilon_{q^p_\mathcal{V}}(g*)$).

### 7.2   A theoretical target risk bound

Using these definitions, we now derive a bound for the target-domain risk of the predictor $g$ (Eq. 1). We do this by an extension of the theory in [1]. Recall that the target risk under consideration is $\epsilon_{q_\mathcal{V}}(g)$. We state the theorem as follows.

**Theorem 1 (Target risk bound).** *Let $\mathcal{H}$ be a hypothesis space and let $g : \mathcal{V} \to \mathcal{Y}$ be a predictor such that $g \in \mathcal{H}$. Given the distributions $p_\mathcal{V}, r_\mathcal{V}, q^s_\mathcal{V}, q^p_\mathcal{V}$, the upper-bound for the target risk $\epsilon_{q_\mathcal{V}}(g)$ is given as,*

$$\epsilon_{q_\mathcal{V}}(g) \leq \epsilon_{p_\mathcal{V}}(g) + \epsilon_{r_\mathcal{V}}(g) + \frac{1}{2}\left\{l_{\mathcal{H}\Delta\mathcal{H}}(q^s_\mathcal{V}, p_\mathcal{V}) + l_{\mathcal{H}\Delta\mathcal{H}}(q^p_\mathcal{V}, r_\mathcal{V})\right\} + \lambda \qquad (8)$$

*where $\lambda = \epsilon_{p_\mathcal{V}}(g*) + \epsilon_{r_\mathcal{V}}(g*) + \epsilon_{q^s_\mathcal{V}}(g*) + \epsilon_{q^p_\mathcal{V}}(g*)$ is the combined risk of the ideal predictor $g* \in \mathcal{H}$.*

**Proof.** By definition, the target-shared and target-private samples correspond to two mutually exclusive sets, and hence, independently contribute to the target risk. Thus, we can decompose the risk as,

$$\underbrace{\epsilon_{q_\mathcal{V}}(g)}_{\text{target-domain risk}} = \underbrace{\epsilon_{q^s_\mathcal{V}}(g)}_{\text{target-shared risk}} + \underbrace{\epsilon_{q^p_\mathcal{V}}(g)}_{\text{target-private risk}} \qquad (9)$$

Consider the first term on the right hand side of Eq. 9, *i.e.* the target-shared risk $\epsilon_{q^s_\mathcal{V}}$. We derive an upper bound for the target-shared risk as follows,

$$\begin{aligned}
\epsilon_{q^s_\mathcal{V}}(g) &= \epsilon_{q^s_\mathcal{V}}(g, h) \\
&\leq \epsilon_{q^s_\mathcal{V}}(g, g*) + \epsilon_{q^s_\mathcal{V}}(g*, h) && \text{using Lemma 1} \\
&= \epsilon_{q^s_\mathcal{V}}(g, g*) - \epsilon_{p_\mathcal{V}}(g, g*) + \epsilon_{p_\mathcal{V}}(g, g*) + \epsilon_{q^s_\mathcal{V}}(g*, h) && \text{introducing } \epsilon_{p_\mathcal{V}}(g, g*) \\
&\leq |\epsilon_{q^s_\mathcal{V}}(g, g*) - \epsilon_{p_\mathcal{V}}(g, g*)| + \epsilon_{p_\mathcal{V}}(g, g*) + \epsilon_{q^s_\mathcal{V}}(g*, h) && \text{modulus operation} \\
&\leq \frac{1}{2}l_{\mathcal{H}\Delta\mathcal{H}}(q^s_\mathcal{V}, p_\mathcal{V}) + \epsilon_{p_\mathcal{V}}(g, g*) + \epsilon_{q^s_\mathcal{V}}(g*, h) && \text{using Definition 3} \\
&\leq \frac{1}{2}l_{\mathcal{H}\Delta\mathcal{H}}(q^s_\mathcal{V}, p_\mathcal{V}) + \epsilon_{p_\mathcal{V}}(g, h) + \epsilon_{p_\mathcal{V}}(g*, h) + \epsilon_{q^s_\mathcal{V}}(g*, h) && \text{using Lemma 1}
\end{aligned}$$

Using the short-hand notations for $\epsilon_{p_\mathcal{V}}(g, h)$, $\epsilon_{p_\mathcal{V}}(g*, h)$ and $\epsilon_{q^s_\mathcal{V}}(g*, h)$ we get,

$$\epsilon_{q^s_\mathcal{V}}(g) \leq \frac{1}{2}l_{\mathcal{H}\Delta\mathcal{H}}(q^s_\mathcal{V}, p_\mathcal{V}) + \epsilon_{p_\mathcal{V}}(g) + \epsilon_{p_\mathcal{V}}(g*) + \epsilon_{q^s_\mathcal{V}}(g*) \qquad (10)$$

In the derivation above, we have the following steps. First, we use the triangle inequality (Lemma 1) to obtain a relation among the disagreement between the

predictor $g$, the ideal predictor $g*$ and the ground-truth labeling function $h$. Then, we introduce the term $\epsilon_{p_\mathcal{V}}(g, g*)$ into the equation (zero-sum), and obtain an upper bound by the property of the modulus operation. At this stage, we analyze the modulus term on the RHS. Recall that the definition of $l_{\mathcal{H}\Delta\mathcal{H}}(d_1, d_2)$ between two distributions $d_1, d_2$ is the supremum of $2|\epsilon_{q_\mathcal{V}^s}(g_1, g_2) - \epsilon_{p_\mathcal{V}}(g_1, g_2)|$ over all $g_1, g_2 \in \mathcal{H}$. Since, the modulus term on RHS ($|\epsilon_{q_\mathcal{V}^s}(g, g*) - \epsilon_{p_\mathcal{V}}(g, g*)|$) is a special case (with $g$ and $g*$), it is upper bounded by $\frac{1}{2}l_{\mathcal{H}\Delta\mathcal{H}}(q_\mathcal{V}^s, p_\mathcal{V})$. Finally, we use the triangle inequality to obtain an upper bound for the term $\epsilon_{p_\mathcal{V}}(g, g*)$. Thus, we obtain the inequality in Eq. 10.

In a similar manner, we can derive an upper bound for the second term on the RHS of Eq. 9, *i.e.* the target-private risk $\epsilon_{q_\mathcal{V}^p}(g)$. Here, we shall consider $l_{\mathcal{H}\Delta\mathcal{H}}(q_\mathcal{V}^p, r_\mathcal{V})$. Now, following the steps used to derive Eq. 10 we get,

$$\begin{aligned}
\epsilon_{q_\mathcal{V}^p}(g) &= \epsilon_{q_\mathcal{V}^p}(g, h) \\
&\leq \epsilon_{q_\mathcal{V}^p}(g, g*) + \epsilon_{q_\mathcal{V}^p}(g*, h) && \textit{using Lemma 1} \\
&= \epsilon_{q_\mathcal{V}^p}(g, g*) - \epsilon_{r_\mathcal{V}}(g, g*) + \epsilon_{r_\mathcal{V}}(g, g*) + \epsilon_{q_\mathcal{V}^p}(g*, h) && \textit{introducing } \epsilon_{r_\mathcal{V}}(g, g*) \\
&\leq |\epsilon_{q_\mathcal{V}^p}(g, g*) - \epsilon_{r_\mathcal{V}}(g, g*)| + \epsilon_{r_\mathcal{V}}(g, g*) + \epsilon_{q_\mathcal{V}^p}(g*, h) && \textit{modulus operation} \\
&\leq \frac{1}{2}l_{\mathcal{H}\Delta\mathcal{H}}(q_\mathcal{V}^p, r_\mathcal{V}) + \epsilon_{r_\mathcal{V}}(g, g*) + \epsilon_{q_\mathcal{V}^p}(g*, h) && \textit{using Definition 3} \\
&\leq \frac{1}{2}l_{\mathcal{H}\Delta\mathcal{H}}(q_\mathcal{V}^p, r_\mathcal{V}) + \epsilon_{r_\mathcal{V}}(g, h) + \epsilon_{r_\mathcal{V}}(g*, h) + \epsilon_{q_\mathcal{V}^p}(g*, h) && \textit{using Lemma 1}
\end{aligned}$$

Thus, the upper bound for target-private risk is,

$$\epsilon_{q_\mathcal{V}^p}(g) \leq \frac{1}{2}l_{\mathcal{H}\Delta\mathcal{H}}(q_\mathcal{V}^p, r_\mathcal{V}) + \epsilon_{r_\mathcal{V}}(g) + \epsilon_{r_\mathcal{V}}(g*) + \epsilon_{q_\mathcal{V}^p}(g*) \tag{11}$$

Finally, we combine the bounds for the target-shared risk (Eq. 10) and target-private risk (Eq. 11), and substitute them in Eq. 9,

$$\begin{aligned}
\epsilon_{q_\mathcal{V}} &= \epsilon_{q_\mathcal{V}^s} + \epsilon_{q_\mathcal{V}^p} \\
&\leq \frac{1}{2}l_{\mathcal{H}\Delta\mathcal{H}}(q_\mathcal{V}^s, p_\mathcal{V}) + \epsilon_{p_\mathcal{V}}(g) + \epsilon_{p_\mathcal{V}}(g*) + \epsilon_{q_\mathcal{V}^s}(g*) \\
&\quad + \frac{1}{2}l_{\mathcal{H}\Delta\mathcal{H}}(q_\mathcal{V}^p, r_\mathcal{V}) + \epsilon_{r_\mathcal{V}}(g) + \epsilon_{r_\mathcal{V}}(g*) + \epsilon_{q_\mathcal{V}^p}(g*)
\end{aligned}$$

Rearranging the terms, we get the target risk bound stated in Theorem 1.

$$\epsilon_{q_\mathcal{V}} \leq \underbrace{\epsilon_{p_\mathcal{V}}(g) + \epsilon_{r_\mathcal{V}}(g)}_{\substack{\text{augmented source} \\ \text{domain risk}}} + \frac{1}{2}\underbrace{\left\{l_{\mathcal{H}\Delta\mathcal{H}}(q_\mathcal{V}^s, p_\mathcal{V}) + l_{\mathcal{H}\Delta\mathcal{H}}(q_\mathcal{V}^p, r_\mathcal{V})\right\}}_{\substack{\text{distribution-shift between target} \\ \text{and augmented source domains}}} \tag{12}$$

$$+ \underbrace{\epsilon_{p_\mathcal{V}}(g*) + \epsilon_{r_\mathcal{V}}(g*) + \epsilon_{q_\mathcal{V}^s}(g*) + \epsilon_{q_\mathcal{V}^p}(g*)}_{\lambda \text{ (minimal combined risk of the ideal predictor)}}$$

□

We have three major terms in Eq. 12 - the augmented source domain risk (source samples augmented with *negative* samples), distribution-shift between the target and the augmented source domains, and the minimal combined risk of the ideal predictor. We will now analyze the significance of this result.

### 7.3   Analysis

We now show that the target risk bound (Eq. 12) is tightened during adaptation.

**Minimizing the distribution-shift.** In the paper, we argue that the adaptation loss $\mathcal{L}_a$ minimizes the distribution-shift in the $\mathcal{V}$-space between the target and the augmented source domain. Here, we describe the process. From the bound in Eq. 12, it is evident that the minimization of the distribution-shift entails the alignment of target-shared and the source distribution, and the alignment of target-private and *negative* sample distribution. Note that, the *negative* samples are essentially used to model the low-source-density region, while the proxy-source samples are used to model the high-source-density region.

In our approach, the loss $\mathcal{L}_{a2}$ pulls the target samples closer to the *guides*. As a result, the target-shared samples are aligned with high source-density region (proxy-source distribution). As we show in the paper in §4.2a, the proxy-source samples are effective in capturing the source-distribution. Therefore, $\mathcal{L}_{a2}$ contributes to the minimization of the term $l_{\mathcal{H}\Delta\mathcal{H}}(q^s_\mathcal{V}, p_\mathcal{V})$. Secondly, $\mathcal{L}_{a1}$ disperses the target clusters away, thereby separating target-private clusters into the low-source-density *negative* region in the $\mathcal{V}$-space. Thus, the shift between the target-private and the low-source-density distribution $l_{\mathcal{H}\Delta\mathcal{H}}(q^p_\mathcal{V}, r_\mathcal{V})$ (modelled by the *negative* samples) is minimized. Overall, the adaptation process enforces the minimization of the distribution-shift between the target and the augmented source domains $l_{\mathcal{H}\Delta\mathcal{H}}(q^s_\mathcal{V}, p_\mathcal{V}) + l_{\mathcal{H}\Delta\mathcal{H}}(q^p_\mathcal{V}, r_\mathcal{V})$.

**Mitigating Catastrophic Forgetting.** We argue that $\mathcal{L}_r = \mathcal{L}_{r1} + \mathcal{L}_{r2}$ mitigates catastrophic forgetting, by preserving a low augmented source risk. Specifically, $\mathcal{L}_{r1}$ is the classification loss applied on the proxy-source samples. The supervision obtained from $\mathcal{L}_{r1}$ ensures that the discriminatory knowledge among the source samples is maintained. This preserves a low source risk $\epsilon_{p_\mathcal{V}}(g)$.

Further, to understand how $\epsilon_{r_\mathcal{V}}(g)$ is minimized, we must study the effect of compact decision boundaries. Intuitively, during the source training (§3.1b of the paper), the $(|\mathcal{C}_s|+1)^{\text{th}}$ logit of $g_s$ is trained to have the highest activation for the *negative* samples, while the source samples are subject to a lower *negative* class activation. This conditioning ensures that in the $\mathcal{U}$-space, whenever a sample falls in the high-source-density region (Fig. 2A of the paper), the highest class logit corresponds to $\mathcal{C}_s$, and vice-versa (if a source class logit is the highest, the sample should pertain to the high-source-density region). Similarly, whenever the source class logits have a lower activation than the *negative* class logit, the sample pertains to the *negative* regime. This property is achieved due to the compact decision boundaries (Fig. 2A of the paper).

During adaptation, by applying the loss $\mathcal{L}_{r1}$ on the proxy-source samples we enforce the logits corresponding to $\mathcal{C}_s$ to have a higher activation than those

corresponding to $\mathcal{C}_t$. Indeed, this ensures that the proxy-source samples fall into the high-source-confidence regions when projected back to the $\mathcal{U}$-space using $f_d$ (see Fig. 2B in the paper). Additionally, $\mathcal{L}_{r2}$ avoids degenerate solutions (such as mode-collapse at the projected $\mathcal{U}$-space), which maintains the property that *negative* samples would be projected to a low-source-density region by $f_d$. Thus, the *negative* samples receive a low source class confidence (alternatively, a higher target-private class confidence from $g$) which is the criterion to detect *negative* samples. Thus, the *negative* sample risk ($\epsilon_{r_\mathcal{V}}(g)$) is minimized. To summarize, the augmented source domain risk $\epsilon_{p_\mathcal{V}}(g) + \epsilon_{r_\mathcal{V}}(g)$ is minimized by $\mathcal{L}_r$.

**Performance of the ideal predictor.** Finally, we also claim that the precision of pseudo-labels for the confident target samples is high. This is verified in Sec. 4. Thus, training the model with $\mathcal{L}_c$ along with $\mathcal{L}_{r1}$ achieves a performance close to the combined risk $\lambda$ of the ideal predictor (referred to as the minimal risk of the optimal joint classifier). This is because, by virtue of $\mathcal{L}_c$ the model is conditioned to minimize the risk on $q_\mathcal{V}^s$ and $q_\mathcal{V}^p$. Further, as argued above, the loss $\mathcal{L}_{r1}$ drives the model to minimize the risk on $p_\mathcal{V}$ and $r_\mathcal{V}$. Note that, if we had enough labeled samples from each of the four distributions, we could learn the ideal predictor $g*$. Nevertheless, using the pseudo-labeled target samples in addition to the proxy-source samples, we can learn a predictor whose performance is close to $g*$.

The observations made above provide a basis for our intuition that the adaptation process tightens the target risk bound (Eq. 12). This enables a superior model upgrade under the CIDA paradigm.

## 8   Discussion on multi-step model upgrade

In this section, we provide details of the multi-step model upgrade experiment. Further, we highlight certain key features of our approach that makes our approach useful in a lifelong learning scenario.

**Approach.** In our approach, a foresighted source-model is trained with the aim of reducing domain and category bias incurred in the given source training dataset. This allows for a source-free model upgrade. We achieve this by modelling Gaussian Prototypes that allows us to enforce the *class separability* objective and to perform *negative* training using samples that fall outside the 3-$\sigma$ confidence region of each class-specific Gaussian Prototype.

We adopt the same strategy during the model upgrade to prepare the target-specific model for a potential future upgrade (*i.e.* a second class-incremental upgrade). Specifically, we add an additional *negative* training loss (similar to $\mathcal{L}_{s2}$) during the first model-upgrade.

**Implementation.** In §4.2f of the paper, we demonstrate CIDA of the (**A**+**D**)-specific model to the **W** domain. During the first upgrade, the feature extractor $f_t$ is adapted to the domain **D**. Simultaneously, we introduce a *negative*-class $(|\mathcal{C}_t'|+1)^{\text{th}}$ to the classifier $g_t$ which is trained to detect *negative* samples in the $\mathcal{V}$-space. These *negative* samples are obtained using the target Gaussian Prototypes

by following a sampling strategy similar to that done in the foresighted source-model training on the domain **A** (*i.e.* we choose samples beyond the 3-$\sigma$ confidence interval of each class-specific target Gaussian Prototype $\mathcal{P}_t^c$).

The model being used for the target domain **D** is $\{f_t, g_t, f_d, g_s\}$. For the second model upgrade, we consider a new domain **W**. In order to adapt to **W**, we initialize the parameters of the **W**-specific feature extractor $f_{t_2}$ with those of $f_t$. For convenience, we denote the **W**-specific semantic space as $\mathcal{V}_2$. The domain projection networks $f_{e_2} : \mathcal{V} \to \mathcal{V}_2$ and $f_{d_2} : \mathcal{V}_2 \to \mathcal{V}$ are pre-trained to a near-identity function. Likewise, we introduce a **W**-specific target-private classifier $g_{t_2}$ which is applied on the $\mathcal{V}_2$-space. Further, the final predictor for **W** is the concatenation of the logits produced by the predictor for **D** and $g_{t_2}$, *i.e.* , $\{\{g_s \circ f_d\}||f_t\}|_{c \in \mathcal{C}_t}$ acting on the $\mathcal{V} - space$ and $g_t$ acting on the $\mathcal{V}_2$-space.

**Suitability for lifelong learning.** One of the key aspects of our proposed framework is the seamless deployment of the target model to multiple steps of CIDA. This is attributed to the following:

- Our approach to tackle CIDA is *source-free* in nature, and does not require the presence of the past source training datasets.
- The adaptation step ensures that the model is ready for a future upgrade on a new domain, without having to be retrained with the previously seen data.
- The domain-specific representation ($\mathcal{V}$-space) allows for a better trade-off between domain-specificity and domain-generalization, where the target-specific model achieves a semantic granularity suitable for the target domain under consideration.
- The domain projection networks $\{f_e, f_d\}$ effectively establish a transit mechanism between two semantically diverse latent spaces, allowing the model to exploit the knowledge of all the previously seen classes.
- The efficacy of the pseudo-labeling process and the selection of confident target samples $\mathcal{B}_t^c$ enhances the reliability of the adaptation process.

We believe that our framework will provide new insights for future studies on lifelong learning in the presence of new domains and tasks.

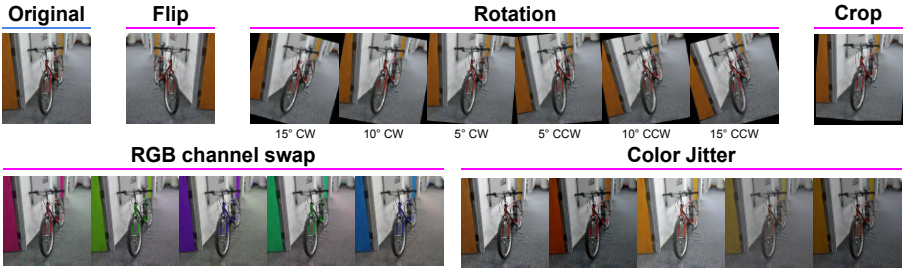

**Fig. 2.** Augmentations applied to the images in **Office** dataset

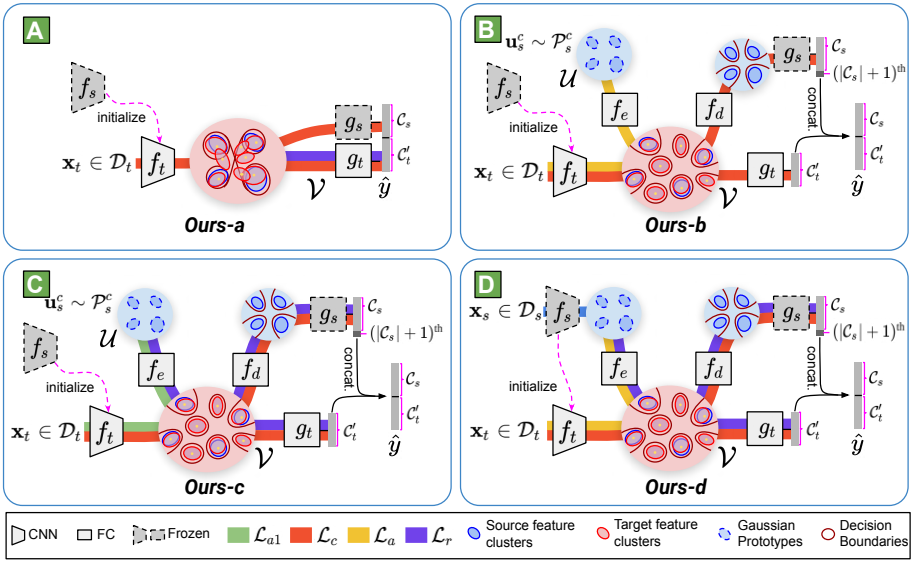

**Fig. 3.** Illustrations of the various baselines of our approach. See §4.2a in the paper for the discussion.

---

**Algorithm 1** GetNegativeSamples

---

1: **require:** Global Gaussian Prototype $\mathcal{P}_s$, Class-specific Gaussian Prototypes $\{\mathcal{P}_s^c : c \in \mathcal{C}_s\}$, number of negative samples $N_{neg}$     /* || denotes append */

2: **initialize:** $\mathbf{U}_n \leftarrow \{\}$, $\mathbf{Y}_n \leftarrow \{\}$

3: **for** $c$ in $\mathcal{C}_s$ **do**   // Obtain 3-$\sigma$ confidence interval thresholds for each class
4:     $\lambda_s^c, \boldsymbol{\nu}_s^c \leftarrow$ largest eigen value and the corresponding eigen vector of $\boldsymbol{\Sigma}_s^c$
5:     $T_s^c \leftarrow \mathcal{P}_s^c(\boldsymbol{\mu}_s^c + 3 \cdot \sqrt{\lambda_s^c} \cdot \boldsymbol{\nu}_s^c)$
6: **end for**

7: **while** $|\mathbf{U}_n| \leq N_{neg}$ **do**   // Obtain *negative* samples
8:     Draw a sample $\mathbf{u}_n \sim \mathcal{P}_s$
9:     $\hat{y}_n \leftarrow \arg\max_c \mathcal{P}_s^c(\mathbf{u}_s)$
10:    **if** $\mathcal{P}_s^{\hat{y}_n}(\mathbf{u}_n) < T_s^{\hat{y}_n}$ **then**
11:        $\mathbf{U}_n \leftarrow \mathbf{U}_n \,||\, \mathbf{u}_n$
12:        $\mathbf{Y}_n \leftarrow \mathbf{Y}_n \,||\, \{unknown\}$
13:    **end if**
14: **end while**

15: **return** $\{\mathbf{U}_n, \mathbf{Y}_n\}$

---

**Algorithm 2** Foresighted Source-model training

1: **require:** Source samples $\mathcal{D}_s$, model parameters $\theta_{f_s}, \theta_{g_s}$, batch size of source samples $N_{src}$ and *negative* samples $N_{neg}$                                 // $\|$ denotes append

2: **repeat** // Pre-train the network for 1 epoch with cross-entropy loss
3:     Obtain a mini-batch of source samples $S_s = \{(\mathbf{x}_s, y_s) \sim p\}$
4:     $\theta_{f_s} \leftarrow \theta_{f_s} + Adam_{f_s}\left(-\nabla_{\theta_{f_s}} \frac{1}{|S_s|} \sum_{(\mathbf{x}_s, y_s) \in S_s} l_{ce}(g_s \circ f_s(\mathbf{x}_s), y_s)\right)$
5: **until** *reached the end of 1 epoch*

    // Calculate Gaussian Prototypes
6: **for** $c \in \mathcal{C}_s$ **do**
7:     $\mathcal{D}_s^c \leftarrow \{(\mathbf{x}_s, y_s) : (\mathbf{x}_s, y_s) \in \mathcal{D}_s \text{ and } y_s = c\}$ // subset of samples in class $c$
8:     $\boldsymbol{\mu}_s^c \leftarrow \text{mean}_{(\mathbf{x}_s, y_s) \in \mathcal{D}_s^c}\left(f_s(\mathbf{x}_s)\right)$ // mean vector
9:     $\boldsymbol{\Sigma}_s^c \leftarrow \text{cov}_{(\mathbf{x}_s, y_s) \in \mathcal{D}_s^c}\left(f_s(\mathbf{x}_s)\right)$ // covariance matrix
10:     $\mathcal{P}_s^c \leftarrow \mathcal{N}(\boldsymbol{\mu}_s^c, \boldsymbol{\Sigma}_s^c)$ // class-specific Gaussian Prototypes
11: **end for**
12: $\boldsymbol{\mu}_s \leftarrow \text{mean}_{(\mathbf{x}_s, y_s) \in \mathcal{D}}\left(f_s(\mathbf{x}_s)\right)$ // mean vector
13: $\boldsymbol{\Sigma}_s \leftarrow \text{cov}_{(\mathbf{x}_s, y_s) \in \mathcal{D}_s}\left(f_s(\mathbf{x}_s)\right)$ // covariance matrix
14: $\mathcal{P}_s \leftarrow \mathcal{N}(\boldsymbol{\mu}_s, \boldsymbol{\Sigma}_s)$ // global Gaussian Prototype

    // Define list of losses and optimizers for alternating minimization
15: **Loss** $\leftarrow [\mathcal{L}_{s1}, \mathcal{L}_{s2}]$
16: **Opt** $\leftarrow [Adam_{\{f_s\}}, Adam_{\{f_s, g_s\}}]$
17: $\boldsymbol{\Theta} \leftarrow [\{\theta_{f_s}\}, \{\theta_{f_s}, \theta_{g_s}\}]$
18: $iter \leftarrow 0$

19: **repeat** // Training loop for source-model
20:     $iter \leftarrow iter + 1$
21:     $cur \leftarrow iter \mod 2$

    // Get a mini-batch of source and *negative* samples
22:     $S_s \leftarrow \{(\mathbf{x}_s, y_s) \sim p\}$ // $|S_s| = N_{src}$
23:     $S_n \leftarrow \text{GetNegativeSamples}(\mathcal{P}_s, \{\mathcal{P}_s^c : c \in \mathcal{C}_s\}, N_{neg})$ // $|S_n| = N_{neg}$
24:     **for** $(\mathbf{x}_s, y_s) \in S_s$ **do**
25:         $\mathbf{y}_s \leftarrow g_s \circ f_s(\mathbf{x}_s)$
26:     **end for**
27:     **for** $(\mathbf{u}_n, y_n) \in S_n$ **do**
28:         $\mathbf{y}_n \leftarrow g_s(\mathbf{u}_n)$
29:     **end for**

    // Optimize the loss **Loss**[*cur*] using the optimizer **Opt**[*cur*]
30:     $\boldsymbol{\Theta}[cur] \leftarrow \boldsymbol{\Theta}[cur] + \mathbf{Opt}[cur]\left(-\nabla_{\boldsymbol{\Theta}[cur]} \mathbf{Loss}[cur]\right)$

31:     **if** *reached the end of an epoch* **then**
32:         Recalculate Gaussian Prototypes following lines 6-14
33:     **end if**
34: **until** *convergence*

---

**Algorithm 3** GetConfidentSamples

1: **require:** Target samples $\mathcal{D}_t$, *guides* $\{\mathbf{v}_g^c : c \in \mathcal{C}_t\}$, percentage of confident samples $n$, model parameters $\theta_{f_t}$            // $\|$ denotes append, $[:]$ denotes indexing

2: **initialize:** $\mathcal{B}_t^c \leftarrow \{\}$  // $\forall c \in \mathcal{C}_t$

3: **for** $\mathbf{x}_t \in \mathcal{D}_t$ **do**
4:     $\mathbf{v}_t \leftarrow f_t(\mathbf{x}_t)$
5:     $d \leftarrow \min_{c \in \mathcal{C}_t} l_2(\mathbf{v}_t, \mathbf{v}_g^c)$  // distance to the nearest *guide*
6:     $k \leftarrow \arg\min_{c \in \mathcal{C}_t} l_2(\mathbf{v}_t, \mathbf{v}_g^c)$  // pseudo-label for the target sample
7: **end for**

8: **for** $c \in \mathcal{C}_t$ **do**
9:     $T \leftarrow$ Target samples with $k = c$ sorted in asc. order of $d$
10:    $b = \frac{n}{100} \cdot |T|$  // number of confident samples for class $c$
11:    $\mathcal{B}_t^c \leftarrow T[0:b]$  // the nearest $b$ samples are the confident target samples for $\mathbf{v}_g^c$
12: **end for**

13: **return** $\{\mathcal{B}_t^c : c \in \mathcal{C}_t\}$

---

---

**Algorithm 4** CIDA on the Target Domain
---
1: **require:** Target samples $\mathcal{D}_t$, Gaussian Prototypes $\mathcal{P}_s^c$, parameters $\theta_{f_s}, \theta_{g_s}, \theta_{f_e}, \theta_{f_d}, \theta_{f_t}, \theta_{g_t}$, percentage of confident samples $n$    // $\|$ denotes append

2: $\theta_{f_t} \leftarrow \theta_{f_s}$ // Initialize $f_t$ from $f_s$
3: **repeat** // Initialize $f_e, f_d$ with near-identity functions
4:     Obtain a mini-batch of proxy-source samples $S = \{\mathbf{u}_s^c \sim \mathcal{P}_s^c : c \in \mathcal{C}_s\}$
5:     $\theta_{f_e} \leftarrow \theta_{f_e} + Adam_{f_e}\big(-\nabla_{\theta_{f_e}} \frac{1}{|S|} \sum_{\mathbf{u}_s^c \in S} l_2(\mathbf{u}_s, f_e(\mathbf{u}_s))^2\big)$
6:     $\theta_{f_d} \leftarrow \theta_{f_d} + Adam_{f_d}\big(-\nabla_{\theta_{f_d}} \frac{1}{|S|} \sum_{\mathbf{u}_s^c \in S} l_2(\mathbf{u}_s, f_d(\mathbf{u}_s))^2\big)$
7: **until** convergence

// Define list of losses and optimizers for round-robin alternating minimization
8: $\mathbf{Loss} \leftarrow [\mathcal{L}_{r1}, \mathcal{L}_{r2}, \mathcal{L}_c, \mathcal{L}_{a1}, \mathcal{L}_{a2}]$
9: $\mathbf{Opt} \leftarrow [Adam_{\{f_e, f_d, g_t\}}, Adam_{\{f_e, f_d\}}, Adam_{\{f_t, g_t\}}, Adam_{\{f_t\}}, Adam_{\{f_e, f_t\}}]$
10: $\boldsymbol{\Theta} \leftarrow [\{\theta_{f_e}, \theta_{f_d}, \theta_{g_t}\}, \{\theta_{f_e}, \theta_{f_d}\}, \{\theta_{f_t}, \theta_{g_t}\}, \{\theta_{f_t}\}, \{\theta_{f_e}, \theta_{f_t}\}]$
11: $iter \leftarrow 0$

12: **repeat** // Training loop for CIDA
13:     $iter \leftarrow iter + 1$
14:     $cur \leftarrow iter \bmod 5$

15:     $\mathbf{v}_g^c \leftarrow f_e(\boldsymbol{\mu}_s^c) \; \forall c \in \mathcal{C}_s, \quad \mathbf{v}_g^c \leftarrow f_t(\tilde{\mathbf{x}}_t^c) \; \forall c \in \mathcal{C}_t'$ // Calculate *guides*

16:     **for** $\mathbf{u}_s^c \in \{\mathbf{u}_s^c \sim \mathcal{P}_s^c : c \in \mathcal{C}_s\}$ **do** // Mini-batch of proxy-source samples
17:         $\mathbf{v}_s^c \leftarrow f_e(\mathbf{u}_s^c); \quad \hat{\mathbf{u}}_s^c \leftarrow f_d(\mathbf{v}_s^c); \quad \hat{\mathbf{y}}_s \leftarrow g_s(\hat{\mathbf{u}}_s^c)|_{c \in \mathcal{C}_s} \;\|\; g_t(\mathbf{v}_s^c)$
18:     **end for**

19:     **for** $\mathbf{x}_t \in \{\mathbf{x}_t \sim q_\mathcal{X}\}$ **do** // Mini-batch of target samples
20:         $\mathbf{v}_t \leftarrow f_t(\mathbf{x}_t); \quad \mathbf{u}_t \leftarrow f_d(\mathbf{v}_t); \quad \hat{\mathbf{y}}_t \leftarrow g_s(\mathbf{u}_t)|_{c \in \mathcal{C}_s} \;\|\; g_t(\mathbf{v}_t)$
21:         $d \leftarrow \min_{c \in \mathcal{C}_t} l_2(\mathbf{v}_t, \mathbf{v}_g^c); \quad k \leftarrow \arg\min_{c \in \mathcal{C}_t} l_2(\mathbf{v}_t, \mathbf{v}_g^c)$ // Pseudo-label
22:     **end for**

23:     **for** $\mathbf{x}_t^c \in \{\mathbf{x}_t^c \sim \mathcal{B}_t^c : c \in \mathcal{C}_t\}$ **do** // Mini-batch of confident target samples
24:         $\mathbf{v}_t^c \leftarrow f_t(\mathbf{x}_t^c), \quad \mathbf{u}_t^c \leftarrow f_d(\mathbf{v}_t^c), \quad \mathbf{y}_t^c \leftarrow g_s(\mathbf{u}_t^c)|_{c \in \mathcal{C}_t} \;\|\; g_t(\mathbf{v}_t^c)$
25:     **end for**

// Optimize the loss **Loss**[*cur*] using the optimizer **Opt**[*cur*]
26:     $\boldsymbol{\Theta}[cur] \leftarrow \boldsymbol{\Theta}[cur] + \mathbf{Opt}[cur]\big(-\nabla_{\boldsymbol{\Theta}[cur]} \mathbf{Loss}[cur]\big)$

27:     **if** *reached the end of an epoch* **then**
28:         Pseudo-label all samples in $\mathcal{D}_t$ using *guides* $\{\mathbf{v}_g^c : c \in \mathcal{C}_t\}$
29:         $\mathcal{P}_t^c \leftarrow$ Gaussian Prototypes obtained using pseudo-labeled target samples
30:         $\mathcal{B}_t^c \leftarrow$ GetConfidentSamples($\mathcal{D}_t, \{\mathbf{v}_g^c : c \in \mathcal{C}_t\}, n$)
31:     **end if**
32: **until** convergence

**Table 2.** The architecture used for training on **Office**, **VisDA** and **Digits** datasets. FC denotes a Fully Connected layer. BN denotes a BatchNorm layer. Conv is Convolutional layer. **Act.** denotes the activation applied at the output of the module.

|       | A. Office and VisDA | | | B. Digits | | |
|-------|---------------------|----------|------|-----------|----------|------|
|       | Module              | Features | Act. | Module    | Features | Act. |
| $f_s$ | ResNet-50 (till avg pool) | 2048 | -    | Conv     | $28 \times 28 \times 64$ | LeakyReLU |
|       | FC                  | 1024     | ELU  | Conv      | $24 \times 24 \times 64$ | LeakyReLU |
|       | BN                  | -        | -    | Conv      | $11 \times 11 \times 128$ | LeakyReLU |
|       | FC                  | 256      | ELU  | Conv      | $5 \times 5 \times 128$ | LeakyReLU |
|       | FC                  | 256      | ELU  | Flatten   | 3200     | -    |
|       | BN                  | -        | -    | FC        | 100      | LeakyReLU |
| $g_s$ | Input               | 256      | -    | Input     | 100      | -    |
|       | FC                  | 64       | ELU  | FC        | 32       | LeakyReLU |
|       | FC                  | $|\mathcal{C}_s|+1$ | - | FC | $|\mathcal{C}_s|+1$ | - |
| $f_t$ | same as $f_s$       | | | same as $f_s$ | | |
| $f_e$ | Input               | 256      | -    | Input     | 100      | -    |
|       | FC                  | 256      | ELU  | FC        | 100      | LeakyReLU |
|       | FC                  | 256      | ELU  | FC        | 100      | LeakyReLU |
|       | FC                  | 256      | ELU  | -         | -        | -    |
| $f_d$ | same as $f_e$       | | | same as $f_e$ | | |
| $g_t$ | Input               | 256      | -    | Input     | 100      | -    |
|       | FC                  | 64       | ELU  | FC        | 32       | LeakyReLU |
|       | FC                  | $|\mathcal{C}'_t|$ | - | FC | $|\mathcal{C}'_t|$ | - |

# References

1. Ben-David, S., Blitzer, J., Crammer, K., Kulesza, A., Pereira, F., Vaughan, J.W.: A theory of learning from different domains. Machine learning **79**(1-2), 151–175 (2010) 6, 7, 8
2. Crammer, K., Kearns, M., Wortman, J.: Learning from multiple sources. Journal of Machine Learning Research **9**(Aug), 1757–1774 (2008) 7
3. Ganin, Y., Ustinova, E., Ajakan, H., Germain, P., Larochelle, H., Laviolette, F., Marchand, M., Lempitsky, V.: Domain-adversarial training of neural networks. JMLR **17**(1), 2096–2030 (2016) 1
4. He, K., Zhang, X., Ren, S., Sun, J.: Deep residual learning for image recognition. In: CVPR (2016) 5
5. LeCun, Y., Bottou, L., Bengio, Y., Haffner, P.: Gradient-based learning applied to document recognition. Proceedings of the IEEE **86**(11), 2278–2324 (1998) 5, 6
6. LeCun, Y., Cortes, C., Burges, C.J.C.: The mnist database of handwritten digits http://yann.lecun.com/exdb/mnist/ 5
7. Liu, H., Cao, Z., Long, M., Wang, J., Yang, Q.: Separate to adapt: Open set domain adaptation via progressive separation. In: CVPR (2019) 1
8. Netzer, Y., Wang, T., Coates, A., Bissacco, A., Wu, B., Ng, A.Y.: Reading digits in natural images with unsupervised feature learning. In: Deep Learning and Unsupervised Feature Learning Workshop at NeurIPS (2011) 5
9. Peng, X., Usman, B., Kaushik, N., Hoffman, J., Wang, D., Saenko, K.: Visda: The visual domain adaptation challenge. arXiv preprint arXiv:1710.06924 (2017) 5
10. Russakovsky, O., Deng, J., Su, H., Krause, J., Satheesh, S., Ma, S., Huang, Z., Karpathy, A., Khosla, A., Bernstein, M., et al.: Imagenet large scale visual recognition challenge. IJCV (2015) 5
11. Saenko, K., Kulis, B., Fritz, M., Darrell, T.: Adapting visual category models to new domains. In: ECCV (2010) 5
12. Saito, K., Yamamoto, S., Ushiku, Y., Harada, T.: Open set domain adaptation by backpropagation. In: ECCV (2018) 1, 6
13. You, K., Long, M., Cao, Z., Wang, J., Jordan, M.I.: Universal domain adaptation. In: CVPR (2019) 1




\bibliographystyle{splncs04}
\bibliography{egbib}

\begin{thebibliography}{10}
\providecommand{\url}[1]{\texttt{#1}}
\providecommand{\urlprefix}{URL }
\providecommand{\doi}[1]{https://doi.org/#1}

\bibitem{OpenSetbaktashmotlagh2018learning}
Baktashmotlagh, M., Faraki, M., Drummond, T., Salzmann, M.: Learning factorized
  representations for open-set domain adaptation. In: ICLR (2019)

\bibitem{ben2010theory}
Ben-David, S., Blitzer, J., Crammer, K., Kulesza, A., Pereira, F., Vaughan,
  J.W.: A theory of learning from different domains. Machine learning
  \textbf{79}(1-2),  151--175 (2010)

\bibitem{ben2007analysis}
Ben-David, S., Blitzer, J., Crammer, K., Pereira, F.: Analysis of
  representations for domain adaptation. In: NeurIPS (2007)

\bibitem{IncrementalCastro_2018_ECCV}
Castro, F.M., Marin-Jimenez, M.J., Guil, N., Schmid, C., Alahari, K.:
  End-to-end incremental learning. In: ECCV (2018)

\bibitem{chang2019dsbn}
Chang, W.G., You, T., Seo, S., Kwak, S., Han, B.: Domain-specific batch
  normalization for unsupervised domain adaptation. In: CVPR (2019)

\bibitem{chapelle2005semi_cluster_assumption}
Chapelle, O., Zien, A.: Semi-supervised classification by low density
  separation. In: AISTATS (2005)

\bibitem{lwm}
Dhar, P., Singh, R.V., Peng, K.C., Wu, Z., Chellappa, R.: Learning without
  memorizing. In: CVPR (2019)

\bibitem{dong2018_oneshot_domainadaptation}
Dong, N., Xing, E.P.: Domain adaption in one-shot learning. In: ECML-PKDD
  (2018)

\bibitem{fort2017gaussian_prototypical_network}
Fort, S.: Gaussian prototypical networks for few-shot learning on omniglot.
  arXiv preprint arXiv:1708.02735  (2017)

\bibitem{NormalAdaptfirstBackprop}
Ganin, Y., Lempitsky, V.: Unsupervised domain adaptation by backpropagation.
  In: ICML (2015)

\bibitem{ganin2016domain}
Ganin, Y., Ustinova, E., Ajakan, H., Germain, P., Larochelle, H., Laviolette,
  F., Marchand, M., Lempitsky, V.: Domain-adversarial training of neural
  networks. JMLR  \textbf{17}(1),  2096--2030 (2016)

\bibitem{Adapt4gong2012geodesic}
Gong, B., Shi, Y., Sha, F., Grauman, K.: Geodesic flow kernel for unsupervised
  domain adaptation. In: CVPR (2012)

\bibitem{goodfellow2013empirical_catastrophic_forgetting}
Goodfellow, I.J., Mirza, M., Xiao, D., Courville, A., Bengio, Y.: An empirical
  investigation of catastrophic forgetting in gradient-based neural networks.
  arXiv preprint arXiv:1312.6211  (2013)

\bibitem{cluster_assumption_grandvalet2005semi}
Grandvalet, Y., Bengio, Y.: Semi-supervised learning by entropy minimization.
  In: NeurIPS (2005)

\bibitem{he2015delving}
He, K., Zhang, X., Ren, S., Sun, J.: Delving deep into rectifiers: Surpassing
  human-level performance on imagenet classification. In: ICCV (2015)

\bibitem{hendrycks2017baseline}
Hendrycks, D., Gimpel, K.: A baseline for detecting misclassified and
  out-of-distribution examples in neural networks. In: ICLR (2017)

\bibitem{ioffe2015batchnorm}
Ioffe, S., Szegedy, C.: Batch normalization: Accelerating deep network training
  by reducing internal covariate shift. In: ICML (2015)

\bibitem{khosla2012undoingbias}
Khosla, A., Zhou, T., Malisiewicz, T., Efros, A.A., Torralba, A.: Undoing the
  damage of dataset bias. In: ECCV (2012)

\bibitem{adam}
Kingma, D.P., Ba, J.L.: Adam: A method for stochastic optimization. In: ICLR
  (2014)

\bibitem{kundu2019gantree}
Kundu, J.N., Gor, M., Agrawal, D., Babu, R.V.: {GAN-Tree}: An incrementally
  learned hierarchical generative framework for multi-modal data distributions.
  In: ICCV (2019)

\bibitem{kundu2019adapt}
Kundu, J.N., Lakkakula, N., Babu, R.V.: {UM-Adapt}: Unsupervised multi-task
  adaptation using adversarial cross-task distillation. In: ICCV (2019)

\bibitem{nath2018adadepth}
Kundu, J.N., Uppala, P.K., Pahuja, A., Babu, R.V.: Adadepth: Unsupervised
  content congruent adaptation for depth estimation. In: CVPR (2018)

\bibitem{usfda}
Kundu, J.N., Venkat, N., M~V, R., Babu, R.V.: Universal source-free domain
  adaptation. In: CVPR (2020)

\bibitem{inheritune}
Kundu, J.N., Venkat, N., Revanur, A., M~V, R., Babu, R.V.: Towards inheritable
  models for open-set domain adaptation. In: CVPR (2020)

\bibitem{kuroki2019unsupervised_discrepancy}
Kuroki, S., Charoenphakdee, N., Bao, H., Honda, J., Sato, I., Sugiyama, M.:
  Unsupervised domain adaptation based on source-guided discrepancy. In: AAAI
  (2019)

\bibitem{lee2013pseudo}
Lee, D.H.: Pseudo-label: The simple and efficient semi-supervised learning
  method for deep neural networks. In: Workshop on Challenges in Representation
  Learning at ICML (2013)

\bibitem{lee2018training_ood}
Lee, K., Lee, H., Lee, K., Shin, J.: Training confidence-calibrated classifiers
  for detecting out-of-distribution samples. In: ICLR (2018)

\bibitem{lwf}
Li, Z., Hoiem, D.: Learning without forgetting. TPAMI  \textbf{40}(12),
  2935--2947 (2017)

\bibitem{sta_open_set}
Liu, H., Cao, Z., Long, M., Wang, J., Yang, Q.: Separate to adapt: Open set
  domain adaptation via progressive separation. In: CVPR (2019)

\bibitem{Adapt2long2015learning}
Long, M., Cao, Y., Wang, J., Jordan, M.I.: Learning transferable features with
  deep adaptation networks. ICML  (2015)

\bibitem{Adapt3long2016unsupervised}
Long, M., Zhu, H., Wang, J., Jordan, M.I.: Unsupervised domain adaptation with
  residual transfer networks. In: NeurIPS (2016)

\bibitem{dfkd}
Lopes, R.G., Fenu, S., Starner, T.: Data-free knowledge distillation for deep
  neural networks. In: LLD Workshop at NeurIPS (2017)

\bibitem{feifei-incremental}
Luo, Z., Zou, Y., Hoffman, J., Fei-Fei, L.F.: Label efficient learning of
  transferable representations across domains and tasks. In: NeurIPS (2017)

\bibitem{zskd}
Nayak, G.K., Mopuri, K.R., Shaj, V., Radhakrishnan, V.B., Chakraborty, A.:
  Zero-shot knowledge distillation in deep networks. In: ICML (2019)

\bibitem{pan2009survey}
Pan, S.J., Yang, Q.: A survey on transfer learning. IEEE Transactions on
  knowledge and data engineering  (2009)

\bibitem{panareda2017open}
Panareda~Busto, P., Gall, J.: Open set domain adaptation. In: ICCV (2017)

\bibitem{mada}
Pei, Z., Cao, Z., Long, M., Wang, J.: Multi-adversarial domain adaptation. In:
  AAAI (2018)

\bibitem{peng2017incrementally}
Peng, H., Li, J., Song, Y., Liu, Y.: Incrementally learning the hierarchical
  softmax function for neural language models. In: AAAI (2017)

\bibitem{visda}
Peng, X., Usman, B., Kaushik, N., Hoffman, J., Wang, D., Saenko, K.: Visda: The
  visual domain adaptation challenge. arXiv preprint arXiv:1710.06924  (2017)

\bibitem{pereyra2017regularizing}
Pereyra, G., Tucker, G., Chorowski, J., Kaiser, {\L}., Hinton, G.: Regularizing
  neural networks by penalizing confident output distributions. In: ICLR (2017)

\bibitem{icarl}
Rebuffi, S.A., Kolesnikov, A., Sperl, G., Lampert, C.H.: icarl: Incremental
  classifier and representation learning. In: CVPR (2017)

\bibitem{ruping2001incremental}
Ruping, S.: Incremental learning with support vector machines. In: ICDM (2001)

\bibitem{office}
Saenko, K., Kulis, B., Fritz, M., Darrell, T.: Adapting visual category models
  to new domains. In: ECCV (2010)

\bibitem{saito2018maximum}
Saito, K., Watanabe, K., Ushiku, Y., Harada, T.: Maximum classifier discrepancy
  for unsupervised domain adaptation. In: CVPR (2018)

\bibitem{Saito_2018_ECCV}
Saito, K., Yamamoto, S., Ushiku, Y., Harada, T.: Open set domain adaptation by
  backpropagation. In: ECCV (2018)

\bibitem{NormalAdaptsankaranarayanan2018generate}
Sankaranarayanan, S., Balaji, Y., Castillo, C.D., Chellappa, R.: Generate to
  adapt: Aligning domains using generative adversarial networks. In: CVPR
  (2018)

\bibitem{shu2019transferable_thuml_weaklysupervised_da}
Shu, Y., Cao, Z., Long, M., Wang, J.: Transferable curriculum for
  weakly-supervised domain adaptation. In: AAAI (2019)

\bibitem{snell2017prototypical}
Snell, J., Swersky, K., Zemel, R.: Prototypical networks for few-shot learning.
  In: NeurIPS (2017)

\bibitem{coral}
Sun, B., Saenko, K.: Deep coral: Correlation alignment for deep domain
  adaptation. In: ECCV Workshops (2016)

\bibitem{torralba2011unbiased}
Torralba, A., Efros, A.A.: Unbiased look at dataset bias. In: CVPR (2011)

\bibitem{tzeng2017adversarial}
Tzeng, E., Hoffman, J., Saenko, K., Darrell, T.: Adversarial discriminative
  domain adaptation. In: CVPR (2017)

\bibitem{NormalAdapttzeng2014deep}
Tzeng, E., Hoffman, J., Zhang, N., Saenko, K., Darrell, T.: Deep domain
  confusion: Maximizing for domain invariance. arXiv preprint arXiv:1412.3474
  (2014)

\bibitem{wu2019largescaleincrementallearning}
Wu, Y., Chen, Y., Wang, L., Ye, Y., Liu, Z., Guo, Y., Fu, Y.: Large scale
  incremental learning. In: CVPR (2019)

\bibitem{UDA_2019_CVPR}
You, K., Long, M., Cao, Z., Wang, J., Jordan, M.I.: Universal domain
  adaptation. In: CVPR (2019)

\bibitem{intrinsic-adversarial}
Zheng, Z., Hong, P.: Robust detection of adversarial attacks by modeling the
  intrinsic properties of deep neural networks. In: NeurIPS (2018)

\end{thebibliography}
\end{document}